%% file: main.tex
\documentclass[journal]{IEEEtran}

\usepackage[T1]{fontenc}
\usepackage{times}
\usepackage{dcolumn}
\usepackage{endnotes}
\usepackage{graphics}
\usepackage{times}
\usepackage{epsfig}
\usepackage{graphicx}
\usepackage{textcomp}
\usepackage{amssymb,amsmath}
\usepackage{balance}

\usepackage{listings}
\usepackage[]{algorithm2e}
\usepackage{flushend}
\usepackage{color,soul}
\usepackage{hyperref}
\usepackage[dvipsnames]{xcolor}
\usepackage{todonotes}

\usepackage{siunitx}
\usepackage{pifont}% http://ctan.org/pkg/pifont
\newcommand{\cmark}{\ding{51}}%
\newcommand{\xmark}{\ding{55}}%

\soulregister\cite7
\soulregister\ref7
\soulregister\pageref7

\begin{document}

%\title{Integer Echo State Networks: Hyperdimensional Reservoir Computing}
\title{Integer Echo State Networks: Efficient Reservoir Computing for Digital Hardware}

% Approach for distributed representation and processing of a sensory data based on principles of hyper-dimensional computing

\author{Denis~Kleyko,
        E.~Paxon~Frady,
        Mansour~Kheffache,
        and~Evgeny~Osipov % <-this % stops a space
\thanks{%Manuscript received on May 1, 2019; 
This work  was  supported in part  by the Swedish Research Council (grant No. 2015-04677). 
The work of DK was supported by the European Union’s Horizon 2020 Research and Innovation Programme under the Marie Skłodowska-Curie Individual Fellowship Grant Agreement 839179 and in part by the DARPA’s VIP (Super-HD Project) and AIE (HyDDENN Project) programs.
%revised May 3, 2018; accepted May 3, 2018. Date of publication May 3, 2018; date of current version May 3, 2018.}
%\thanks{This work was supported by the Swedish Research Council (grant no. 2015-04677) and Systems on Nanoscale Information fabriCs (SONIC), one of the six SRC STARnet Centers, sponsored by MARCO and DARPA. DK also acknowledges Stiftelsen Seth M Kempes Stipendiefond for partially funding his research visit to UC Berkeley.
}
\thanks{\mbox{*}D. Kleyko is with the Redwood Center for Theoretical Neuroscience at the University of California, Berkeley, CA 94720, USA and also with Intelligent Systems Lab at Research Institutes of Sweden, 164 40 Kista, Sweden. \mbox{E-mail}: \mbox{denis.kleyko@ri.se}}% <-this % stops a space 
%\thanks{E. P. Frady is with Neuromorphic Computing Lab, Intel Labs and also with the Redwood Center for Theoretical Neuroscience at the University of California, Berkeley, CA 94720, USA. \mbox{E-mail}: \mbox{epaxon@berkeley.edu}
\thanks{E. P. Frady is with the Redwood Center for Theoretical Neuroscience at the University of California, Berkeley, CA 94720, USA. \mbox{E-mail}: \mbox{epaxon@berkeley.edu}
}% <-this % stops a space
\thanks{M. Kheffache is with Netlight Consulting AB, 111 53 Stockholm, Sweden. \mbox{E-mail}: \mbox{mansour.kheffache@netlight.com}}% <-this % stops a space 
\thanks{E. Osipov is with the Department of Computer  Science Electrical and Space Engineering, Lule\aa{} University of Technology, 971 87 Lule\aa{}, Sweden. \mbox{E-mail}: \mbox{evgeny.osipov@ltu.se} }% <-this % stops a space
 }% <-this % stops a space

\markboth{}%
{Kleyko \MakeLowercase{\textit{et al.}}: Integer Echo State Networks}

\maketitle

% I would change the first sentence (the rest is fine) to:
%
\begin{abstract}

%We propose an integer approximation of Echo State Networks (ESN) based on the
%mathematics of hyperdimensional computing.
%The reservoir of the proposed Integer Echo State Network (intESN) contains only \textit{n}-bits integers and replaces the recurrent matrix multiply %with an efficient cyclic shift operation. Such an architecture results in dramatic improvements in memory footprint and computational efficiency, with minimal performance loss.
%Our architecture naturally supports the usage of the trained reservoir in  symbolic processing tasks of analogy making and logical inference. 

We propose an approximation of Echo State Networks (ESN) that can be efficiently implemented on digital hardware based on the mathematics of hyperdimensional computing. The reservoir of the proposed integer Echo State Network (intESN) is a vector containing only n-bits integers (where n<8 is normally sufficient for a satisfactory performance). The recurrent matrix multiplication is replaced with an efficient cyclic shift operation. The proposed intESN approach is verified with typical tasks in reservoir computing: memorizing of a sequence of inputs; classifying time-series; learning dynamic processes. Such architecture results in dramatic improvements in memory footprint and computational efficiency, with minimal performance loss. 
%{\color{red}
The experiments on a field-programmable gate array confirm that the proposed intESN approach is much more energy efficient than the conventional ESN.
%}
\end{abstract}


\input{tex/introduction} 
\input{tex/Related}

\input{tex/intESN}

\input{tex/performance_evaluation}

\input{tex/fpga}

\input{tex/Discussion}
\input{tex/Conclusions}

\bibliography{bica}

\end{document}

%% file: tex/Introduction.tex
%\vspace*{-0.6cm}
\section{Introduction}
\label{sect:intro}
%\vspace*{-0.3cm}

%\todo[inline, color=olive]{
%DK: Adapt to the presence of hardware experiments \\
%Introduce new results on modeling chaotic systems
%}

%\hl{
Recent work in Reservoir Computing \cite{RC09, Sussillo2009} (RC) illustrates how a Recurrent Neural Network with fixed connectivity can memorize and generate complex spatio-temporal sequences. RC has been shown to be a powerful tool for modeling and predicting dynamic systems, both living \cite{ESN11NIPS} and technical \cite{ESN04, RCnature11}. 
%{\color{red}
Recently, it has been shown in~\cite{CHAOS18} that RC is able to predict large chaotic systems. 
%}

Recent work on feed-forward networks shows that the binarization of filters in Convolutional Neural Networks can lead to enormous gains in memory and computational efficiency \cite{ECCV16}. Reducing the memory allocated to each neuron or synapse from a 32-bit float to a few bits or binary saves computation with minimal loss in performance (see, e.g., \cite{BinNN, QuanNN}). The increase in efficiency broadens the range of applications for these networks. 

This article addresses two important research directions in RC: training reservoir networks and implementing networks efficiently. We discovered several direct functional similarities between the operations in RC and those of hyperdimensional computing (HDC) \cite{Frady17}. HDC \cite{Kanerva09} or more generally Vector Symbolic Architectures \cite{Gayler2003} are frameworks for neural symbolic representation, computation, and analogical reasoning. The distinction from the traditional computing is that all entities (objects, phonemes, symbols) are represented by random vectors of very high dimensionality -- several thousand dimensions. Complex data structures and analogical reasoning are implemented by simple arithmetical operations (binding, addition/bundling, and permutation) and a well-defined similarity metric \cite{Kanerva09}.
%}
Specifically, RC and HDC are connected by the following core principles:
~
\begin{itemize}
  \item Random projections of input values onto a reservoir (which in essence is a high-dimensional vector) matches random HDC representations stored in a superposition; 
  \item The update of the reservoir by a random recurrent connection matrix is similar to HDC binding/permutation operation;
  \item The nonlinearity of the reservoir can be approximated with the thresholded addition of integers in HDC. 
\end{itemize}
~
We exploited these findings to design integer Echo State Networks (intESN), which perform like Echo State Networks (ESN) but with smaller memory footprint and computational cost. 

In the proposed architecture, the reservoir of the network contains only constrained integers for each neuron, reducing the memory of each neuron from a 32-bit float to only a few bits. The recurrent matrix multiply update is replaced by a permutation (or even a cyclic shift), which results in the dramatic boosting of the computational efficiency. We validate the architecture on several tasks common in the RC literature. All examples demonstrate satisfactory approximation of performance of the conventional ESN;
%{\color{red}
while the implementation on field-programmable gate array confirms the amenability of intESN for digital hardware.

The article is structured as follows. Background and related work are presented
in Section \ref{sect:related}. The main contribution -- integer Echo State
Networks are described in Section \ref{sect:intesn}. The performance evaluation
follows in Section \ref{sect:perf}. 
%{\color{red}
Section \ref{sect:fpga} presents the experiments on digital hardware. 
%}
Sections \ref{sect:dis} and \ref{sect:conclusions}
present a discussion and conclusions.

% Both areas HDC and RC have features in common. In particular,  both are using
% distributed representation of data in high-dimensional spaces.  For example,
% ESNs use random projections \cite{RachRaPr14} to map data into 
% high-dimensional spaces. HDC usually operates with symbolic data from  a
% finite alphabet. Symbols are assigned with unique random high-dimensional vectors.
% 
% This article cross-fertilizes both areas and proposes an architecture for ESNs (intESNs) where the reservoir operates with integer numbers in the limited range. 
% The advantage of the proposed architecture is that in contrast to the standard ESN it does not require floating point operations and storage in the reservoir but the procedure for training the readout layer of neurons is the same as in the ESN. The reservoir nonlinearity operation and the mapping of data into the reservoir are taken from the HDC. This development is in line with the current research attempts to improve the computational efficiency of ANNs   by quantizing or even binarizing the training process \cite{BinNN, QuanNN, ECCV16}. 
% Also, a combination of both approaches makes a step further towards addressing the problem of neural-symbolic integration \cite{NSIsur}.
% The proposed architecture is compared with the standard ESN \cite{ESN04} in terms of its memory as well for the task of predicting dynamic system on the examples of a sinusoid function and Mackey-Glass equation. 

%% file: tex/related.tex
\section{Background and related work}
\label{sect:related}

There are many practical tasks that require the history of inputs to be solved.
In the area of artificial neural networks (ANN), such tasks require working memory. This could be implemented by recurrent connections between neurons of an RNN. 
%Such networks are called recurrent neural networks (RNNs). 
Training RNNs is much harder than that of feed-forward ANNs (FFNNs) due to vanishing gradient problem \cite{Bengio94}.

The challenge of training RNNs was addressed from two approaches. 
One approach eliminates the vanishing gradient problem through neurons with special memory gates \cite{LSTM97}. 
%First, the vanishing gradient problem can be eliminated through neurons with special memory gates, as it is done in Long Short-Term Memory \cite{LSTM97}.
Another approach is to reformulate the training process by learning only connections to
the last readout layer while keeping the other connections fixed.
This approach originally appeared in two similar architectures: Liquid State Machines
\cite{LSM02} and ESNs \cite{ESN03}, now referred to as reservoir computing\cite{RC09}.
%\hl{
%Despite, the simplicity of the readout layer training this process has its own nuances as the trained layer can be unstable. This problem was recently covered in \cite{Kudithipudi2018}. 
%}

It is interesting to note, that similar ideas were conceived in the area of
FFNNs, which can be seen as an RNN without memory, and are known under the name
of Extreme Learning Machines (ELMs) \cite{ELM06}.
ELMs are used to solve various machine learning problems including
classification, clustering, and regression \cite{ELM15}.

%\hl{
%The RC is a  powerful tool for modeling and predicting dynamic systems both living \cite{ESN11NIPS} and technical \cite{ESN04, RCnature11}  systems. 
%The RC is useful for  modeling and predicting dynamic systems. Another important application of the RC is a classification of time-series. 
Important applications of RC are the modeling and predicting of complex dynamic systems. 
%{\color{brown}
Generating and predicting chaotic systems was an important use-case from the beginning~\cite{Jaeger2001}, for example, ESNs were used for chaotic time-series from low-order aberrations caused by turbulence~\cite{Weddell2008}.
A thorough study on emulating chaotic systems was recently presented in~\cite{Antonik2017}.
%}
It was also shown that ESNs can be used for forecasting of Electroencephalography signals and for solving classification problems in the context of Brain-Computer Interfaces \cite{ESNEEG}.
There are different classification strategies and readout methods when performing classification of time-series with ESN. 
In \cite{ComparisonReadOut} three classification strategies and three readout methods were explored under the conditions that testing data is purposefully polluted with noise. 
Interestingly, different readout methods are preferable in different noise conditions. 
Recent work in \cite{Bianchi2018} also studied classification of multivariate time-series with ESN using standard benchmarking datasets. 
The work covered several advanced approaches, which extend the conventional ESN architecture, for generating a representation of a time-series in a reservoir. 
%}

Another recent research area is binary RC with cellular automata (CARC) which started as a interdisciplinary research within three areas: cellular automata, RC, and HDC. CARC was initially explored in \cite{Yilmaz15a}  for projecting binarized features into high-dimensional space.  Further in \cite{ISBI}, it was applied for modality classification of medical images. The usage of CARC for symbolic reasoning is explored in \cite{Yilmaz15b}. The memory characteristics of a reservoir formed by CARC are presented in  \cite{CAHD17}. Work \cite{RCELMCA17} proposed the usage of coupled cellular automata in CARC.    
%\hl{
Examples of recent RC developments also include advanced architectures such as Laplacian ESN~\cite{Han2018}, learning of reservoir's size and topology~\cite{Qiao2017}, new tools for investigating reservoir dynamics~\cite{BianchiTNNLS18} and determining its edge of criticality~\cite{Livi2018}.
%}

%{\color{brown}
The design of ESNs has been an important research area (see, e.g.,~\cite{Ozturk2007, Busing2010, Strauss2012}). 
One important aspect of the design is, of course, the choice of network's parameters for a given task.
Another important aspect considered in this study is the computational complexity. 
One of the ways of reducing computational costs would be to use quantized reservoir states. 
It was explored in Fractal Prediction Machines~\cite{Tino00} and Neural Prediction Machines~\cite{Tino04, Tino07} RC models. 
Another way of reducing computational costs involves modifications of network's connectivity structure. 
In this respect, the approach which is ideologically closest to our intESN, was presented in~\cite{MinESN}. 
The authors demonstrated that a simple cycle reservoir (referred to as the ring-based ESN) can be used to achieve a performance similar to the conventional ESN. 
Similar conclusions about the ring-based ESN were obtained in~\cite{Strauss2012} when studying different design strategies for reservoir connection matrices in four typical RC tasks. 
While the ring-based ESN explored reservoir update solution, which is similar to one of our optimizations, the technical side is very different from our approach as intESN strives at using only integers as neurons activation values.
%}

%{\color{brown}
While in this article the main focus is on reservoir states comprised of integers only, it is worth mentioning related works considering the general problem of reducing the computational complexity of the conventional ESN. 
For example, several optimizations were used in~\cite{ESNAnomaly09} in order to deploy an ESN on a resource-constrained device for anomaly detection.
These optimizations include sparse matrix algebra via compressed row storage for weights of connections between between the input layer neurons and the reservoir; single floating point precision; and an activation function, which resembles $\tanh()$ function but has lower complexity. 
Similarly, in works~\cite{Bacciu2013, Bacciu2014} ESNs were used in the context of user movements prediction on resource-constrained devices, therefore, the authors studied how the parameters of the network will affect computational costs and task performance.
In particular, the varied parameters were sparsity of reservoir connection matrix, number of bits per weight, number of neurons in reservoir.
%}

% Overview of the major theories
 
\input{ESN_overview}

\input{hd_theory}

%% file: ESN_overview.tex
\subsection{Echo State Networks}
\label{sect:esn}
% 

% \begin{figure}[hbt]
% \minipage{0.49\textwidth}
%   \includegraphics[width=\linewidth]{img/ESN_new}
%   \caption{Architecture of the conventional Echo State Network.}
% \label{fig:esn}
% \endminipage\hfill
% \minipage{0.49\textwidth}%
%   \includegraphics[width=\linewidth]{img/HD_ESN_new}
%   \caption{Architecture of the Integer Echo State Network.}
% \label{fig:intesn}
% \endminipage
% \end{figure}

\begin{figure}[tb]%[!ht]%[t!]
\centering
\includegraphics[width=1.0\columnwidth]{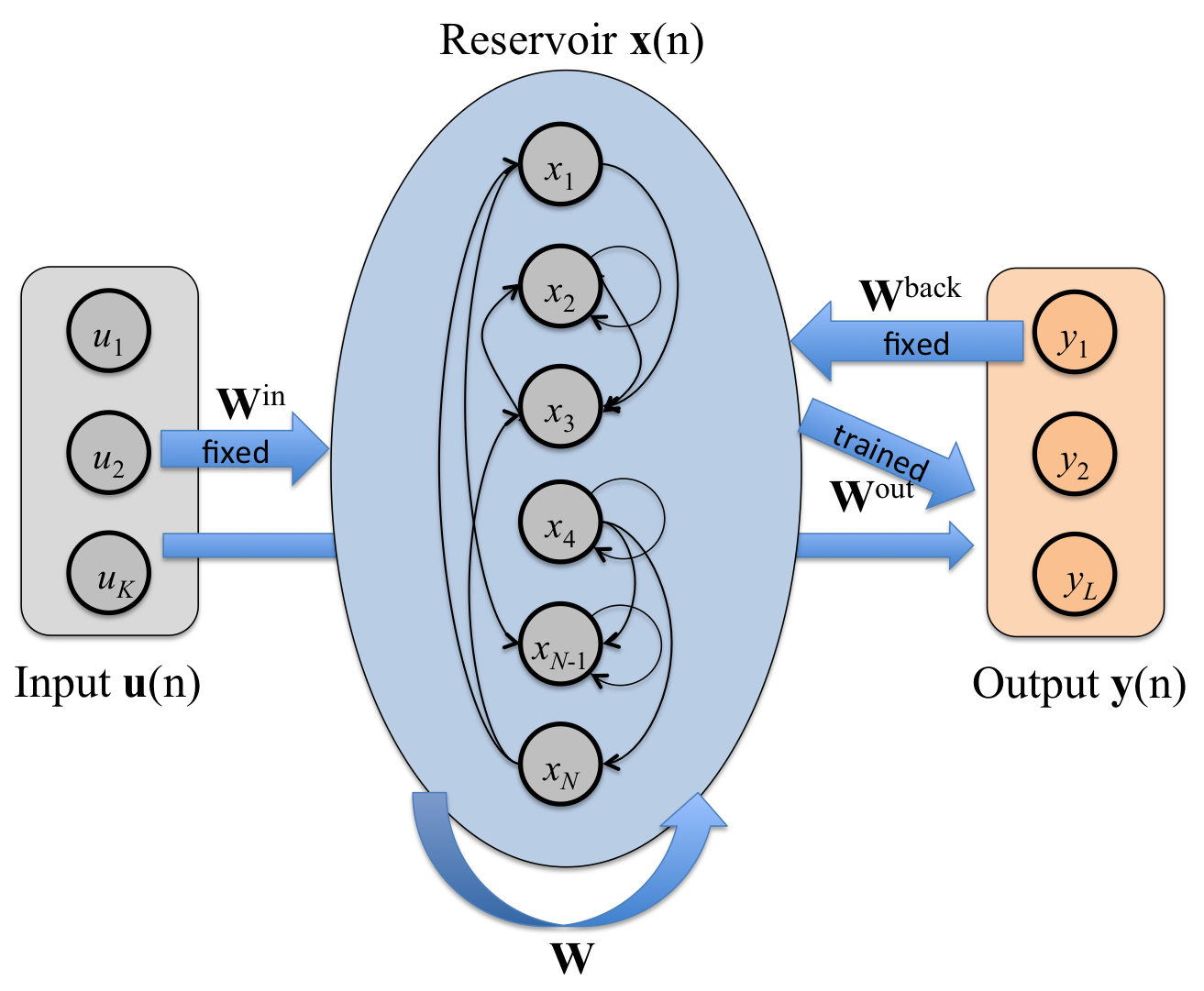}
\caption{Architecture of the conventional Echo State Network.}
\label{fig:esn}
%\vspace*{-0.5cm}
\end{figure}

%\hl{
This subsection summarizes the functionality of the conventional ESN, it follows the description in \cite{ESNtut12} for a special case of leaky integration when $\alpha=1$\footnote{For the detailed tutorial on ESNs diligent readers are referred to \cite{ESNtut12}.}.
%}
Fig.~\ref{fig:esn} depicts the architectural design of the conventional ESN, which includes three layers of neurons. The input layer with $K$ neurons represents the current value of input signal denoted as $\textbf{u}(n)$. The output layer ($L$ neurons) produces the output of the network (denoted as $\textbf{y}(n)$) during the operating phase. The reservoir is the hidden layer of the network with $N$ neurons, with the state of the reservoir at time $n$ denoted as $\textbf{x}(n)$. 

In general, the connectivity of ESN is described by four matrices. $\textbf{W}^{\text{in}}$ describes connections between the input layer neurons and the reservoir, and $\textbf{W}^{\text{back}}$ does the same for the output layer. Both matrices project the current input and output to the reservoir.   
The memory in ESN is due to the recurrent connections between neurons in the reservoir, which are described in the reservoir matrix $\textbf{W}$.  
Finally, the matrix of readout connections $\textbf{W}^{\text{out}}$ transforms the current activity levels in the input layer and reservoir ($\textbf{u}(n)$ and $\textbf{x}(n)$, respectively) into the network's output $\textbf{y}(n)$.  
~
~
Note that three matrices ($\textbf{W}^{\text{in}}$, $\textbf{W}^{\text{back}}$, and $\textbf{W}$) are randomly generated at the network initialization and stay fixed during the network's lifetime. Thus, the training process is focused on learning the readout matrix $\textbf{W}^{\text{out}}$. There are no strict restrictions for the generation of projection matrices   
$\textbf{W}^{\text{in}}$ and $\textbf{W}^{\text{back}}$. They are usually randomly drawn from either normal or uniform distributions and scaled as shown below. The reservoir connection matrix, however, is restricted to posses the echo state property. This property is achieved when the  spectral radius of the matrix  $\textbf{W}$ is less or equal than one. 
%\hl{
For example, $\textbf{W}$ can be generated from a normal distribution and then normalized by its maximal eigenvalue.
Unless otherwise stated, in this article an orthogonal matrix was used as the reservoir connection matrix; such a matrix was formed by applying QR decomposition to a random matrix generated from the standard normal distribution. Also, $\textbf{W}$ can be scaled by a feedback strength parameter, see (\ref{eq:esnres}).
%}

The update of the network's reservoir at time $n$ is described by the following equation:
~
\begin{equation}
\textbf{x}(n)=\tanh(\rho\textbf{W}\textbf{x}(n-1)+\beta\textbf{W}^{\text{in}}\textbf{u}(n)+\beta\textbf{W}^{\text{back}}\textbf{y}(n-1)),
\label{eq:esnres}
 \end{equation}
%\hl{
where  $\beta$ and $\rho$ denote projection gain and the feedback strength, respectively. Note that it is assumed that the spectral radius of the reservoir connection matrix $\textbf{W}$ is one.
%}
Note also that at each time step neurons in the reservoir apply $\tanh()$ as the activation function.
The nonlinearity prevents the network from exploding by restricting the range of possible values from -1 to 1. The activity in the output layer is calculated as:  
~
\begin{equation}
\hat{\textbf{y}}(n)=g(\textbf{W}^{\text{out}}[\textbf{x}(n);\textbf{u}(n)]),
\label{eq:esny}
 \end{equation}
where the semicolon denotes concatenation of two vectors and $g()$ the activation function of the output neurons, for example, linear or  Winner-take-all.  
~
\subsubsection{Training process}
\label{sect:training}

This article only considers training with supervised-learning when the network is provided with the ground truth desired output at each update step. The reservoir states $\textbf{x}(n)$ are collected together with the ground truth $\textbf{y}(n)$ for each training step. The weights of the output layer connections are acquired by solving the regression problem which minimizes the mean square error between predictions (\ref{eq:esny}) and the ground truth.
%For example, $\textbf{W}^{\text{out}}$ can be calculated using the pseudo-inverse matrix operation (adopted in this article). 
While this article does not focus on the readout training task, it should be noted that there are many alternatives reported in the literature including the usage of regression with regularization, online update rules, etc. \cite{ESNtut12}.

%% file: hd_theory.tex
\subsection{Fundamentals of hyperdimensional computing}
\label{sect:sparse}

In a localist representation, which is used in all modern digital
computers, a group of bits is needed in its entirety to interpret a representation.  
In HDC, all entities
(objects, phonemes, symbols, items) are represented by vectors of very high
dimensionality -- thousands of bits. The information is spread out in a {\it distributed representation}, which contrary to the
localist representations, any subset of the bits can be interpreted. 
Computing with distributed representations utilizes statistical
properties of vector spaces with very high dimensionality, which allow for approximate, noise-tolerant, highly parallel computations.
%The results are evaluated only in terms of a well-defined similarity metric.
{\it Item memory} (also referred to as {\it clean-up} memory) is needed to recover
composite representations assigned to complex concepts. There are several flavors
of HDC with distributed representations, differentiated by the
random distribution of vector elements, which can be real numbers \cite{PlateTr,
Gallant, MAP, Gallant2016}, complex  numbers \cite{PlateBook}, binary numbers
\cite{Kanerva09, Rachkovskij2001}, or bipolar \cite{Gallant, HD_ICRC16}. 

We rely on the mathematics of HDC with  
{\it bipolar distributed representations} to develop intESN.
%\subsection{Computing with dense bipolar distributed representations} 
Kanerva \cite{Kanerva09} proposed the use of distributed representations comprising $N=10, 000$
{\it binary} elements (referred to as HD vectors). The values of each element of an HD vector are independent
equally probable, hence they are also called dense distributed representations.
Similarity between two binary HD vectors is characterized by  Hamming distance,
which (for two vectors) measures the number of elements in which they differ. 
In very high dimensions  Hamming distances (normalized by the dimensionality $N$)
 between any arbitrary chosen HD vector and all other vectors in the
HD space are concentrated around 0.5. Interested readers are referred to
\cite{Kanerva09} and \cite{KanervaBook} for comprehensive
analysis of probabilistic properties of the high-dimensional
representational space.

The binary HD vectors can be equivalently mapped to
the case of bipolar representations, i.e., where each vector's element is
encoded as  ``-1'' or ``+1''. This definition is sometimes more convenient for
purely computational reasons. 
%Bipolar dense distributed representations ,however, possess a set of distinctive properties. 
The distance metric for the bipolar case is a dot product:
~
\begin{equation}
\text{dist} = \textbf{x}^\top \textbf{y}
\end{equation}
%\begin{equation}
  %   dist=|\textbf{x} \wedge \textbf{y}|_1.
 %\end{equation}
~
%\hl{
Basic symbols in HDC are referred to as  atomic HD vectors. They are generated randomly and independently,  and due to
high dimensionality will be nearly orthogonal with very high probability, i.e., similarity
(dot product) between such HD vectors is approximately 0.
An ordered sequence of symbols can be encoded into a composite HD vector
%, resembling a reservoir in RC, 
using the atomic HD vectors, the permutation (e.g., cyclic shift as a special case of permutation) and bundling operations.
This vector encodes the entire sequence history in the composite HD vector and resembles a neural reservoir. 
%}

%\hl{
Normally in HDC, the recovery of component atomic HD vectors from a composite HD vector is performed by finding the most similar vectors stored in the item memory\footnote{It is not common to do such decoding in RC. Normally, in the scope of RC a readout matrix is learned. In this article, we follow this standard RC approach to extracting information back from a reservoir.}. 
However, as more vectors are bundled together there is more interference noise and the likelihood of recovering the correct atomic HD vector declines. 

Our recent work \cite{Frady17} reveals the impact of interference noise, and shows that different flavors of HDC have universal memory capacity. Thus, the different flavors of HDC can be interchanged without affecting performance. From these insights, we are able to design much more efficient networks for reservoir computing for digital hardware. 
%}

%% file: tex/intESN.tex
%\vspace*{-0.6cm}
\section{Integer Echo State Networks}
\label{sect:intesn}
%\vspace*{-0.3cm}
% 
% \begin{figure}[!ht]%[t!]
% \centering
% \includegraphics[width=0.7\columnwidth]{img/HD_ESN}
% \caption{Architecture of the Integer Echo State Network.}
% \label{fig:intesn}
% %\vspace*{-0.5cm}
% \end{figure}

% \begin{figure}[hbt]
% \minipage{0.49\textwidth}
%   \includegraphics[width=\linewidth]{img/HD_ESN}
%   \caption{Architecture of the Integer Echo State Network.}
% \label{fig:intesn}
% \endminipage\hfill
% \minipage{0.49\textwidth}%
%   \includegraphics[width=\linewidth]{img/Discretization}
%   \caption{Quantization and discretization of a continuous signal.}
% \label{fig:quantization}
% \endminipage
% \end{figure}

\begin{figure}[tb]%[!ht]%[t!]
\centering
\includegraphics[width=1.0\columnwidth]{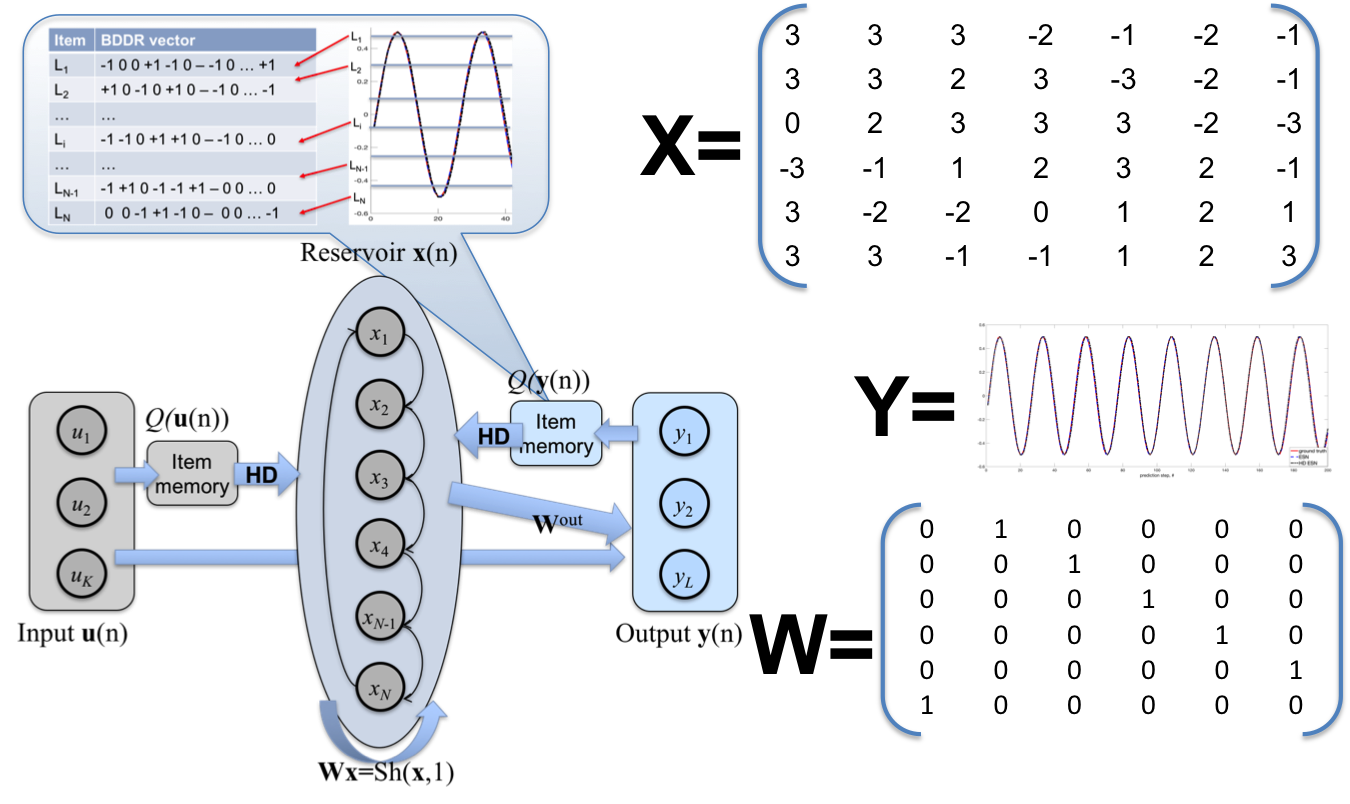}
\caption{Architecture of the proposed integer Echo State Network.}
\label{fig:intesn}
%\vspace*{-0.5cm}
\end{figure}

This section presents the main contribution of the article -- an architecture for integer Echo State Network. 
The architecture is illustrated in Fig.~\ref{fig:intesn}. The proposed intESN is structurally identical to the the conventional ESN (see Fig.~\ref{fig:esn}) with three layers of neurons: input ($\textbf{u}(n)$, $K$ neurons), output ($\textbf{y}(n)$, $L$ neurons), and reservoir ($\textbf{x}(n)$, $N$ neurons). It is important to note from the beginning that training the readout matrix $\textbf{W}^{\text{out}}$  for intESN is the same as for the conventional ESN (Section \ref{sect:training}). 

However, other components of intESN differs from the conventional ESN.  First,
activations of input and output layers are projected into the  reservoir in the
form of bipolar HD vectors \cite{MAP} of size $N$  (denoted as
$\textbf{u}^{\text{HD}}(n)$ and $\textbf{y}^{\text{HD}}(n)$). 
 % you said this in the previous section:
% Such projection is achieved by mapping activations into high-dimensional vectors.  According to the principles of HDC, each ``symbol''  is assigned with a random high-dimensional vector used as representation in HDC.  The correspondences are stored in the so-called item memory (Fig.~\ref{fig:intesn}),  which given the symbol issues the corresponding high-dimensional vector  (denoted as HD in the figure). 
For problems where input and output data are described by finite
alphabets and each symbol can be treated independently,  the mapping to $N$-dimensional space is achieved by simply assigning a random bipolar HD vector
to each symbol in the alphabet and storing them in the item memory \cite{Kanerva09, Kleyko2015}.  In the
case with continuous data (e.g., real numbers), we quantized the continuous values
into a finite alphabet.
%A symbol in such alphabet is a quantization level.  
The quantization scheme (denoted as $Q$) and the granularity of the
quantization are problem dependent. Additionally, when there is a need to
preserve similarity between quantization levels, distance preserving mapping
schemes are applied (see, e.g., \cite{Scalarencoding, Widdows15}), which can
preserve, for example, linear or nonlinear similarity between levels. An example
of a discretization and quantization of a continuous signal as well as its
HD vectors in the item memory is illustrated in Fig.~\ref{fig:intesn}. 
%\hl{
Continuous values can be also represented in HD vectors by varying their density. For a recent overview of several mapping approaches readers are referred to \cite{TNNLS18}. Also, an example of applying such mapping is presented in Section~\ref{sect:perf:analog}.
%}
%Thus after choosing the mapping scheme activations of neurons are projected into bipolar HD vector and this vector is added to reservoir.
Another feature of intESN is the way the recurrence in the reservoir is implemented. Rather than a matrix multiply, recurrence is implemented via the permutation of the reservoir vector. Note that permutation of a vector can be described in matrix form, which can play the role of $\textbf{W}$ in intESN. Note that the spectral radius of this matrix equals one.
However, an efficient implementation of permutation can be achieved for a special case -- cyclic shift (denoted as $\text{Sh}()$). 
%{\color{red}
It is important to note that we have shown in~\cite{Frady17} that the recurrent weight matrix $\textbf{W}$ creates key-value pairs of the input data.
Note that $\textbf{W}$ is chosen randomly and kept fixed, and this always leads to the same properties.
Moreover, there is no advantage of the fully connected random recurrent weight matrix over the simple cyclic shift operation for storing the input history.
Thus, the use of the cyclic shift in place of a random  recurrent weight matrix does not limit intESN's ability to produce linearly separable representations. 
%}
Fig.~\ref{fig:intesn} shows the recurrent connections of neurons in a reservoir with recurrence by cyclic shift of one position. In this case, vector-matrix multiplication $ \textbf{W} \textbf{x}(n) $ is equivalent to $ \text{Sh}(\textbf{x}(n),1)$.

% In the ESN the reservoir matrix is a random matrix  which posses specific properties as described in the previous section.
%\hl{
Finally, to keep the integer values of neurons, intESN uses different nonlinear activation function for the reservoir -- clipping (\ref{eq:clipping}).
%}
Note that the simplest bundling operation is an elementwise addition. However, when using the elementwise
addition, the activity of a reservoir (i.e., a composite HD vector) is no longer bipolar. From the implementation
point of view, it is practical to keep the values of the elements of the HD vector in the limited range using  a threshold value (denoted as $\kappa$). % and make it a configurable parameter. The bounding operation is called {\it clipping}. The clipping is done as follows:
~
\begin{equation}
f_\kappa (x) = 
\begin{cases}
-\kappa & x \leq -\kappa \\
x & -\kappa < x < \kappa \\
\kappa & x \geq \kappa
\end{cases}
\label{eq:clipping}
\end{equation}
~
The clipping threshold $\kappa$ is regulating nonlinear behavior of the reservoir and limiting the range of activation values. Note that in intESN the reservoir is updated only with integer bipolar vectors, and after clipping the values of neurons are still integers in the range between $-\kappa$ and $\kappa$. Thus, each neuron can be represented using only $\log_2(2\kappa+1)$ bits of memory. For example, when $\kappa=7$, there are fifteen unique values of a neuron, which can be stored with just four bits. 
We have also shown recently that the usage of the clipping might be beneficial when implementing resource-efficient alternatives of Self-Organizing Maps~\cite{intSOM}.

Summarizing the aforementioned differences, the update of intESN is described as: 
~
\begin{equation}
\textbf{x}(n)= f_\kappa (\text{Sh}(\textbf{x}(n-1),1)+\textbf{u}^{\text{HD}}(n)+\textbf{y}^{\text{HD}}(n-1)).
\label{eq:intesnres}
 \end{equation}

%% file: tex/performance_evaluation.tex
\section{Performance evaluation}
\label{sect:perf}

%\hl{
In this section, the proposed intESN approach is verified and compared to the conventional ESN and the ring-based ESN~\cite{MinESN} on a set of typical RC tasks.
In particular, three aspects are evaluated: short-term memory, classification of time-series, and modeling of dynamic processes.
Short-term memories are compared using the trajectory association task
\cite{PlateBook}, introduced in the area of holographic reduced representations
\cite{PlateTr}. Additionally, an approach for storing and decoding analog values using intESN is demonstrated on image patches. 
Classification of time-series is studied using the standard datasets from UCI and UCR.
Modeling of dynamic processes is tested on two typical cases. First, the task of learning a simple sinusoidal function is considered. Next, networks are trained to reproduce a complex dynamical system produced by a Mackey-Glass series.
Unless otherwise stated, ridge regression (the regularization coefficient is denoted as $\lambda$) with the  Moore-Penrose pseudo-inverse was used to learn the
readout matrix $\textbf{W}^{\text{out}}$. Values of input neurons $\textbf{u}(n)$ were not used for training the readout in any of the experiments below. 
%}
%Note about studies of capacity for both RC and HPC
%Moore-Penrose pseudo-inverse of matrix

\subsection{Short-term memory}
\subsubsection{Sequence recall task}
\label{sect:perf:trajectory}

%While we leave formal analysis of the capacity for future work 

The sequence recall task includes  two stages: memorization and recall.
At the memorization stage,  a network continuously stores a sequence of tokens 
(e.g., letters, phonemes, etc).
The number of unique tokens is denoted as $D$ ($D=27$ in the experiments), and one token is presented as input each timestep.
At the recall stage, the network uses the content of its reservoir to
retrieve the token stored $d$ steps ago, where $d$ denotes delay. In the
experiments, the range of delay varied between 0 and 15.

\begin{figure}[tb]%[!ht]%[t!]
\centering
\includegraphics[width=1.0\linewidth]{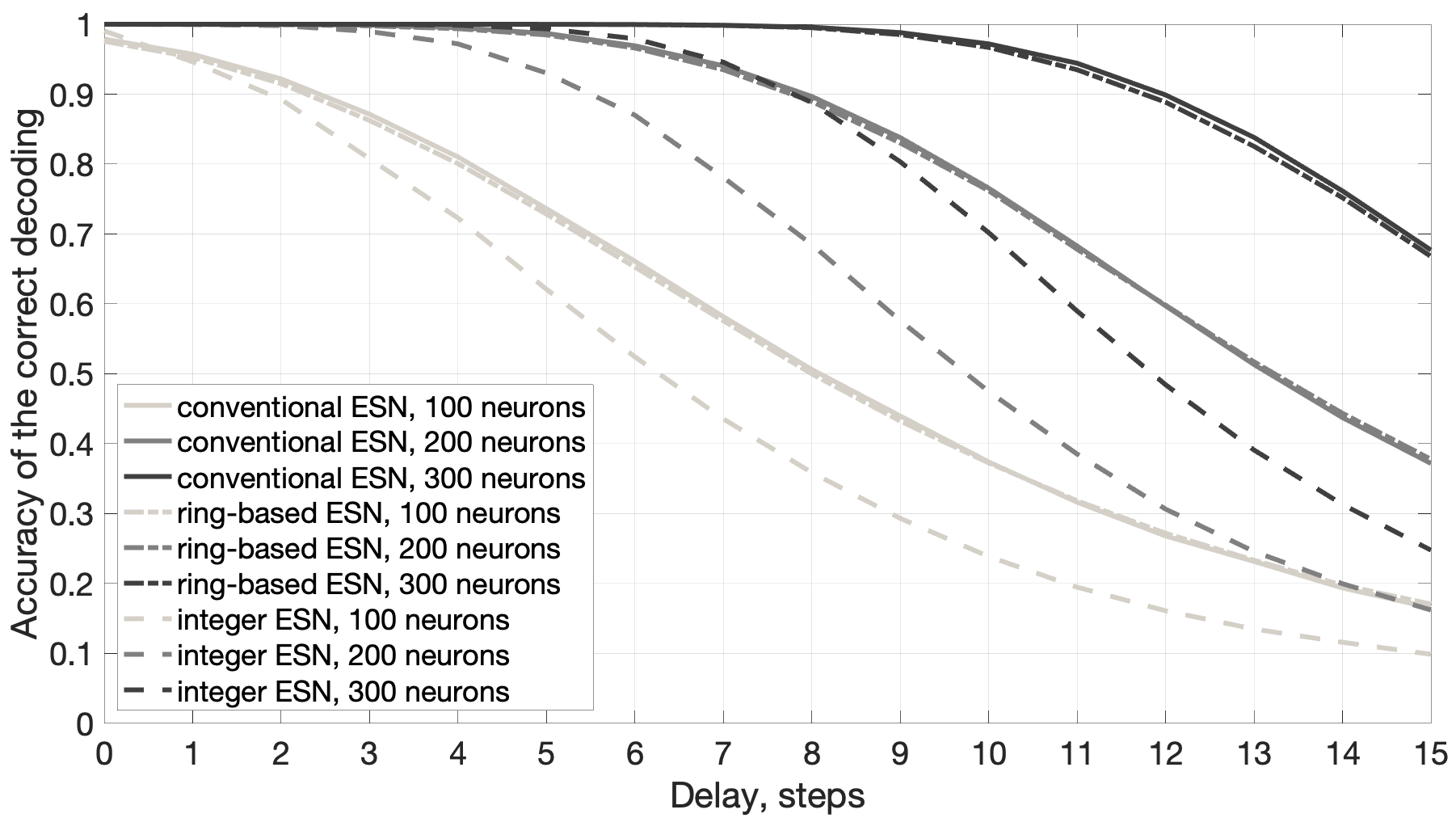}
\caption{The accuracy of the  correct decoding of tokens for the conventional ESN, ring-based ESN, and integer ESN for
three different values of $N$.
}
\label{fig:memory}
%\vspace*{-0.5cm}
\end{figure}

For the conventional and ring-based ESNs, the  dictionary of tokens was represented by a one-hot encoding, i.e.
the number of input layer neurons was set to the  size of the dictionary
$K=D=27$. The same encoding scheme was adopted for the output layer, $L=27$.
%\hl{
The input vector  was projected to the reservoir by the projection matrix
$\textbf{W}^{in}$ where each entry was independently generated from the uniform distribution in the
range $[-1,1]$, the projection gain was set to $\beta=0.1$.  
The reservoir connection matrix $\textbf{W}$ for the conventional ESN  was first generated from the standard normal distribution and then orthogonalized. 
The reservoir connection matrix $\textbf{W}$ for the ring-based ESN  was generated as a permutation matrix. 
The feedback strength of both reservoir connection matrices was set to $\rho=0.94$. 
%}

For intESN, the item memory was populated with $D$ random high-dimensional
bipolar vectors.  The threshold for the clipping function was set to $\kappa=3$. The
output layer was the same as in ESN with $L=27$ and one-hot encoding of tokens.
%{\color{brown}
It is worth noting that $\rho$ and $\kappa$ were chosen in such a way that the accuracy curves would resemble each other as close as possible. 
The diligent readers are kindly referred to the Supplementary materials (Fig. S.1) where for the case $N=200$ the curves for the range of $\rho$ and $\kappa$ values are presented. 
%}

For each value of of the delay $d$ a readout matrix $\textbf{W}^{\text{out}}$  was trained, producing 16 matrices in total. 
The training sequence presented 2000 random tokens to the network, and only the last 1500 steps were used to compute the readout matrices. The regularization parameter for ridge regression was set to $\lambda=0$.
The training sequence of tokens delayed by the particular $d$ was used as the ground truth for the the activations of the output layer.  
During the operating phase,  both the inclusion of a new token into the reservoir and the recall of the delayed token from the reservoir 
were simultaneous.  Experiments were performed for three different sizes of the reservoir: $N=100$, $N=200$, and $N=300$.

\begin{figure}[tb]%[!ht]%[t!]
\centering
\includegraphics[width=1.0\linewidth]{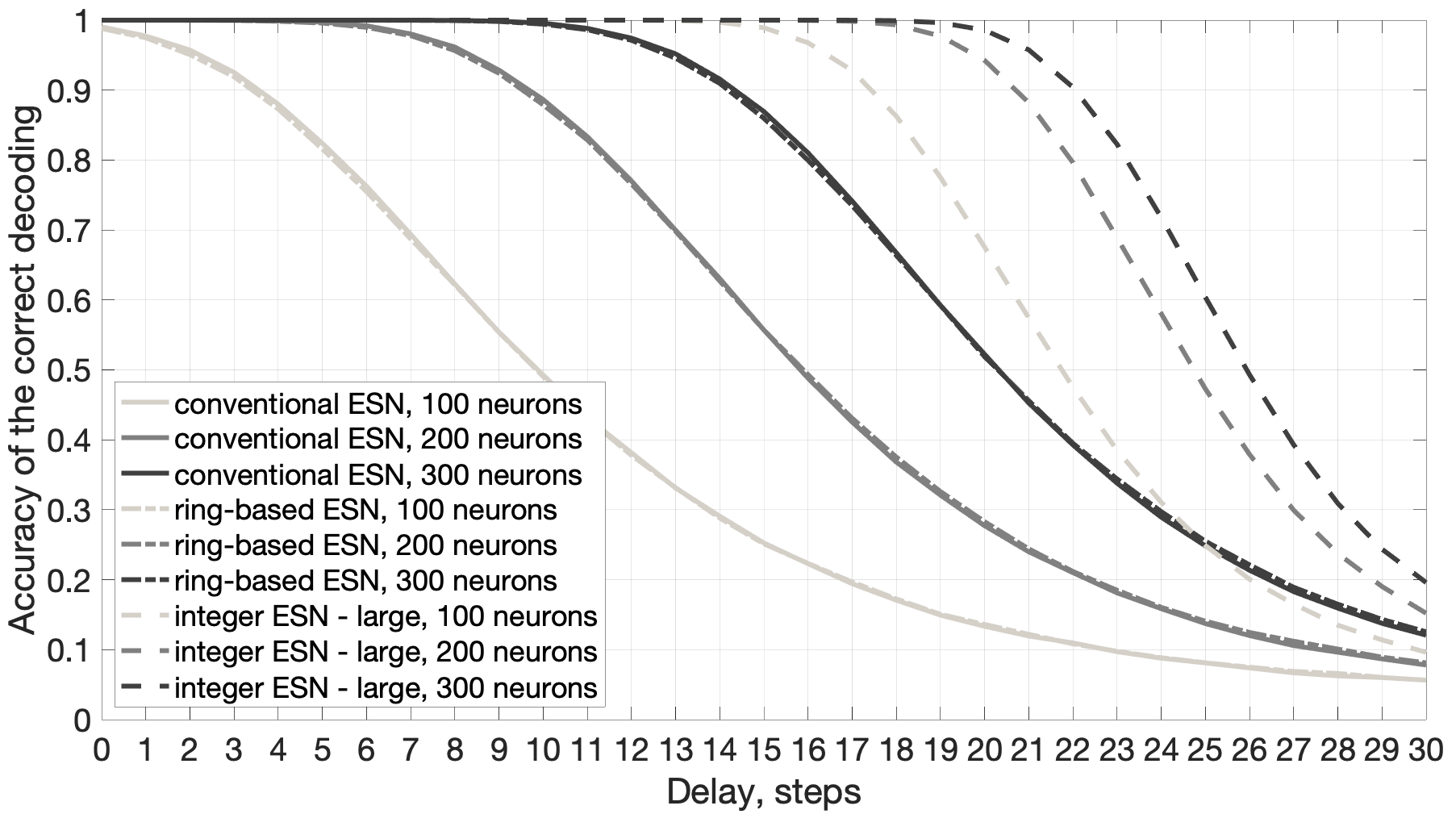}
\caption{The accuracy of the  correct decoding of tokens for the conventional ESN, ring-based ESN, and integer ESN for
three different values of $N$. ``intESN-large'' refers to the fact that the number of neurons in intESN was equivalent to the memory footprint required by ESN for the stated number of neurons.  
}
\label{fig:memory:long}
%\vspace*{-0.5cm}
\end{figure}

The memory capacity of the network is characterized by the 
accuracy of the correct decoding of tokens for different values of the delay. 
%{\color{brown}
Fig.~\ref{fig:memory} depicts the accuracy for all networks conventional ESN (solid lines), ring-based ESN (dash-dotted line)
and intESN (dashed lines). The capacities of all the networks grow with the
increased number of neurons in the reservoir.  
Since the capacities of the conventional ESN and the ring-based ESN are almost identical, which is in line with~\cite{Strauss2012}, for the rest of this subsection we assume both of these networks when using the term ESN.  
%}

\begin{figure*}[tb]%[h]
\center{
\begin{minipage}[h]{0.8\linewidth}
\center{\includegraphics[width=1.0\linewidth]{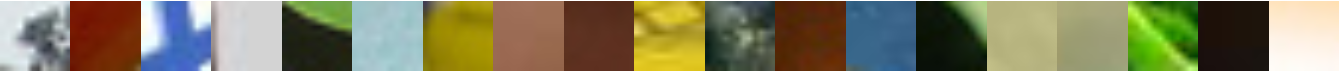} }
\end{minipage}
\vfill
\begin{minipage}[h]{0.8\linewidth}
\center{\includegraphics[width=1.0\linewidth]{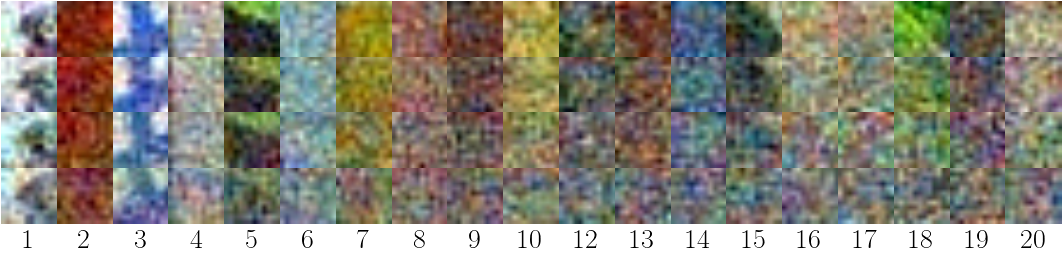}}
\end{minipage}
\caption{
%\hl{
An example of image patches decoded from an intESN. Top row represents the original images stored in the reservoir.   
Other rows depict the patches reconstructed from intESN for different reservoir sizes and clipping thresholds.
%}
}
}
\label{fig:images}
\end{figure*}

%\hl{
The capacities of ESN and intESN  are comparable for small $d$, i.e., for the most recent tokens. 
For the increased delays the curves featured slightly different behaviors. 
With increase of the value of $d$ the performance of intESN started to decline faster compared to ESN.
%For all values of $N$ the accuracy of  intESN starts to deviate from 100\% earlier than that of  ESN. Also, intESN features slightly steeper decay than ESN. 
Eventually, all curves converge to the value of the random guess which equals $1/D$. 
%Moreover, it is possible to characterize the information capacity using a single number -- the amount of information which can be decoded from the reservoir. 
Moreover, the information capacity of a network is characterized by the amount of the decoded from the reservoir information.
This amount is determined using the amount of information per token ($\log_2D$), the probability of correctly decoding a token at each delay value, and the concept of mutual information. We calculated the amount of information for all networks in Fig.~\ref{fig:memory} in the considered delay range. For 100 neurons intESN preserved 19.3\% less information, for 200 and 300 neurons 21.7\% less.
%}

%\hl{
%On the other hand, intESN with $\kappa=3$ requires only 3-bit per neuron. It is assumed that one ESN neuron requires 32-bit then if the number of neurons in %intESN is increased ten times the reservoir memory footprints of two networks are going to be comparable. The results for this case are presented in Fig.~\ref{fig:memory:long} (the training sequence was prolonged to 9000 random tokens). In such setting intESN has clearly higher information capacity. In particular, %for ESN memory footprint with 100 neurons the decoded amount of information has increased 2.2 times while for 200 and 300 neurons it increased 1.6 and 1.3 times %respectively. 
%}

%\hl{
These results highlight a very important trade-off: the performance versus a complexity of implementation. While
the performance of intESN is somewhat poorer in this task, one has to bear in mind its memory footprint. With the
clipping threshold $\kappa=3$ only 3-bit are needed to represent the state of a neuron compared to 32-bit
per neuron (the size of type float) in ESN. 
%{\color{red}
In other words, intESN allowed lowering the memory footprint of the reservoir by an order of magnitude by sacrificing only a fraction of the performance with respect to the information capacity. 
Thus, we conjecture that some reduction in the performance for ten folds
memory footprint reduction is an acceptable price in applications on resource-constrained computing
devices.
%}
On the other hand, we can check the performance of the networks with equal memory
footprints. 
For this we increased the number of neurons in intESN so that the total memory consumed by
the reservoir with the same clipping threshold $\kappa=3$ would match that of the conventional or ring-based ESN. 
This network is denoted as ``intESN-large''.
%{\color{red}
Since $\kappa=3$ requires only 3-bit, in order to get the memory footprint corresponding to ESN, intESN could use more than ten times more neurons. 
Thus, the memory footprint of intESN with $1000$ neurons corresponds to ESN with $100$ neurons; while intESN with $2000$ and $3000$  neurons correspond to ESN with $200$ and $300$ neurons respectively.
%}
 The results for this case are presented in Fig.~\ref{fig:memory:long} (the training sequence was prolonged to
9000 random tokens). With such settings intESN-large has clearly higher information capacity. In
particular, for ESN memory footprint with 100 neurons the decoded amount of information has increased
2.2 times while for 200 and 300 neurons it increased 1.6 and 1.3 times respectively.
%}
%{\color{brown}
It is important to note, however, that while the memory consumed by the reservoir of intESN-large was comparable to the corresponding ESN, the readout matrix for intESN-large was larger and more computationally demanding than the ESN readout matrix since the size of a readout matrix is proportional to the number of neurons in the reservoir. 
%}

%ESN: 100 - 36.9;  200 - 58.6; 300 - 70.5; 
%intESN: 100 - 29.8;  200 - 45.9; 300 - 55.2; 
%Long training.
%ESN: 100 - 46.9;  200 - 74.9; 300 - 95.43; 
%intESN: 100 - 103.3;  200 - 117.4; 300 - 122.8; 

\subsubsection{Storage of analog values in intESN}
\label{sect:perf:analog}

%\hl{
This subsection presents the feasibility of storing analog values in intESN using image patches as a showcase. 
%{\color{red}
It is important to emphasize that this subsection  does not go into detailed comparisons with other methods as the main purpose here is 
the principal demonstration of the possibilities of storing continuous data in reservoirs consisting of integers in a limited range. 
In other words, with this showcase, we are aiming at demonstrating the feasibility of using integer approximation of neuron states in intESN to work with analog representations.
%}
A value of a pixel (in an RGB channel) can be treated as an analog value in the range between 0 and 1. 
For each pixel it is possible to generate a unique bipolar HD vector. 
%The pixel's value is encoded by multiplying all elements of the HD vector by that value. 
The typical approach to encode an analog value is to multiply all elements of the HD vector by that value.
The whole sequence is then represented using the bundling operation on all scaled HD vectors. The result of bundling can be used as an input to a reservoir. However, the resultant composite HD vector will not be in the integer range anymore. 
%This could be addressed in using scaling via sparsity. 
We address this problem by using sparsity.
Instead of scaling elements of an HD vector, we propose to randomly set the fraction of elements of the HD vector to zeros, i.e., the HD vector will become ternary. The proportion of zero elements is determined by the pixel's analog value. Pixels with values close to zero will have very sparse HD vectors while pixels with values close to one will have dense HD vectors, but all entries will always be
[-1, 0, or +1].
The result of bundling of such HD vectors (i.e., HD vector for an image) will still have integer values. Such representational scheme allows keeping integer values in the reservoir but it still can effectively store analog values.
%}

The examples of results are presented in Fig.~\ref{fig:images}. Top row depicts original images stored in the reservoir. The other rows depict images reconstructed from the reservoir. The following parameters of intESN were used (top to bottom): $N=64000$, $\kappa=11$; $N=32000$, $\kappa=8$;  $N=16000$, $\kappa=6$;  $N=8000$, $\kappa=4$. The values of $\kappa$ were optimized for a particular $N$. Columns correspond to the delay values (i.e., how many steps ago an image was stored in the reservoir) as in the previous experiment. 
As one would anticipate, the quality of the reconstructed images is improving for larger reservoir sizes. At the same time, the quality of the reconstructed images is deteriorating for larger delay values, i.e., the worst quality of the reconstructed image could be observed in the bottom right corner while the best reconstruction is located in the top left corner.
Nevertheless, the main observation for this experiment is that it is possible to project analog values into the reservoir with integer values using the mapping via varying sparsity and then retrieve the values from the reservoir.
Moreover, we have shown recently~\cite{intRVFL} that the mapping via varying sparsity could even be helpful when solving classification problems with a feed-forward variant of the ESN.

\begin{table}[tb]%[ht]
\renewcommand{\arraystretch}{1.3}
\caption{Details of datasets for time-series classification.\label{tab:datasets}
\vspace{-2mm}}
    % {\scriptsize
    \begin{center}
    \begin{tabular}{|c|c|c|c|c|}\hline
      	\multicolumn{5}{|c|}{\textbf{Univariate datasets from UCR}} \\ \hline\hline
        \textbf{Name} & \textit{\textbf{\#V}} & \textbf{Train} & \textbf{Test} & \textit{\textbf{\#C}} \\ \hline
        Swedish Leaf 	& 1	& 500 	& 625 	& 15 \\ \hline  
 	Distal Phalanx & 1	& 139 & 400 &	3 \\ \hline
	ECG 		& 1 	& 100 & 100 &	2 \\ \hline	
        Wafer  		& 1 	& 1000	 & 6164 &	2\\ \hline\hline       
        \multicolumn{5}{|c|}{\textbf{Multivariate datasets from UCI}} \\ \hline\hline
	 Character Trajectories & 3 		& 300 & 2558 & 20  \\\hline
	 Spoken Arabic Digit 	 & 13 	& 6600 & 2200 & 10  \\\hline
        Japanese Vowels 	& 12 	& 270 & 370 & 9 \\ \hline    	
    \end{tabular}
    \end{center}
%  }
%\vspace{-5mm}
\end{table}

\subsection{Classification of time-series}

%\hl{
In this section, ESN (conventional and ring-based) and intESN networks are compared in terms of classification accuracy obtained on standard time-series datasets. 
Following \cite{Bianchi2018} we used several (four) univariate datasets from UCR\footnote{UCR. Time Series Classification Archive [online], 2018. -- Available online: \url{https://www.cs.ucr.edu/\%7Eeamonn/time\_series\_data\_2018/}.}  and several (three) multivariate datasets from UCI\footnote{UCI. Machine Learning Repository [online], 2019. -- Available online: \url{http://archive.ics.uci.edu/ml/datasets.html}.}.
Details of datasets are presented in Table~\ref{tab:datasets}. 
For each dataset, the table includes the name, number of variables (\#\textit{V}), number of classes (\#\textit{C}), and the number of examples in training and testing datasets.  
%}

%\hl{
Configurations of the networks were kept fixed for all datasets. 
%{\color{brown}
In fact, the configuration of the conventional and ring-based ESNs were set in accordance to~\cite{Bianchi2018}: 
reservoir size was set to $N=800$, projection gain was set to $\beta=0.25$, the feedback strength was set to $\rho=0.99$. 
The regularization parameter for the ridge regression was set to $\lambda=1.0$.
The intESN was also trained with the same $\lambda$.
%}
The clipping threshold for the intESN  was set to $\kappa=7$. 
%{\color{red}
Also for intESN the quantized values of time-series were mapped to bipolar vectors using scatter codes \cite{TNNLS18, scatter}. 
The input signal \textbf{u}(n) was quantized as:
~
\begin{equation}
\textbf{u}(n)_q= \lfloor 200\textbf{u}(n) \rceil /200,
\label{eq:quan:timeser}
 \end{equation}
~
where $\lfloor * \rceil$ denotes rounding to the the closest integer. 
%}
Two sizes of intESN's reservoir were used. The first size corresponded to the size of the conventional and ring-based ESNs, i.e., $N=800$. The second size (``intESN-large'') corresponded to the same memory footprint\footnote{Except for the Japanese Vowels dataset where such reservoir size seemed to significantly overfit the training data. In that case, the number of neurons was increased twice.} required for ESN reservoir assuming that one ESN neuron requires 32-bit while one intESN neuron requires 4-bit (when $\kappa=7$). 
%{\color{red}
Thus, ``intESN-large'' had  $N=6400$ neurons. 
%}
%}

%\hl{
The output layers of networks were representing one-hot encodings of classes in a dataset, i.e., for the particular dataset $L=$\#\textit{C} of that dataset. 
The readout layers of all networks were trained using time-series from a training dataset in the so-called endpoints mode \cite{ComparisonReadOut} when only final temporal reservoir states for each time-series  are used for training a single readout matrix. 
%}

\begin{figure}[tb]%[!ht]%[t!]
\centering
\includegraphics[width=1.0\linewidth]{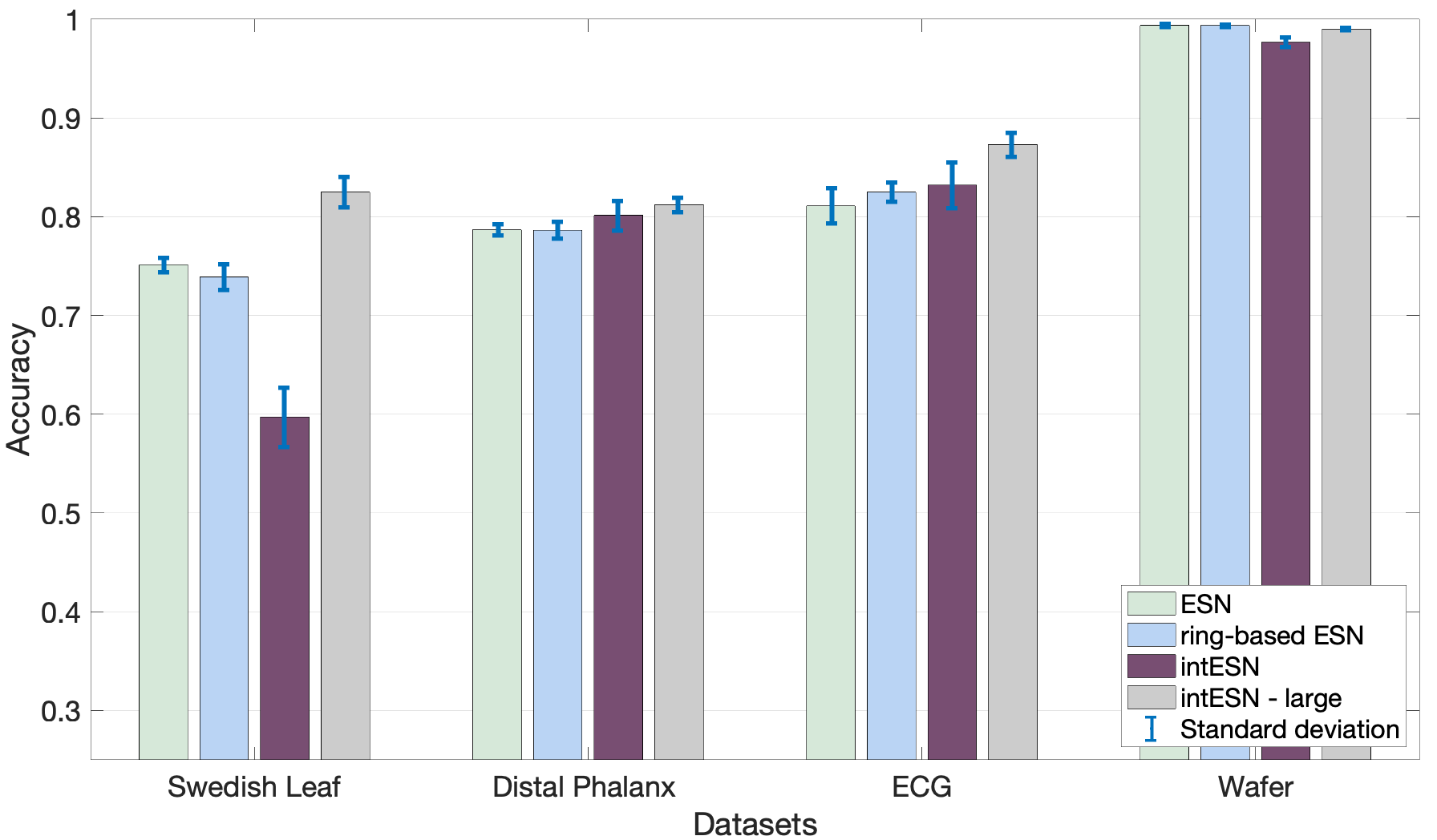}
\caption{The classification accuracy for univariate datasets from UCR. Bars depict mean values, lines depict standard deviations.  Bars denoted as ``ESN'' and ``intESN'' had the same number of neurons in their reservoirs while for ``intESN-large'' the number of neurons corresponded to ESN's memory footprint. 
}
\label{fig:univar}
\end{figure}

\begin{figure}[tb]%[!ht]%[t!]
\centering
\includegraphics[width=1.0\linewidth]{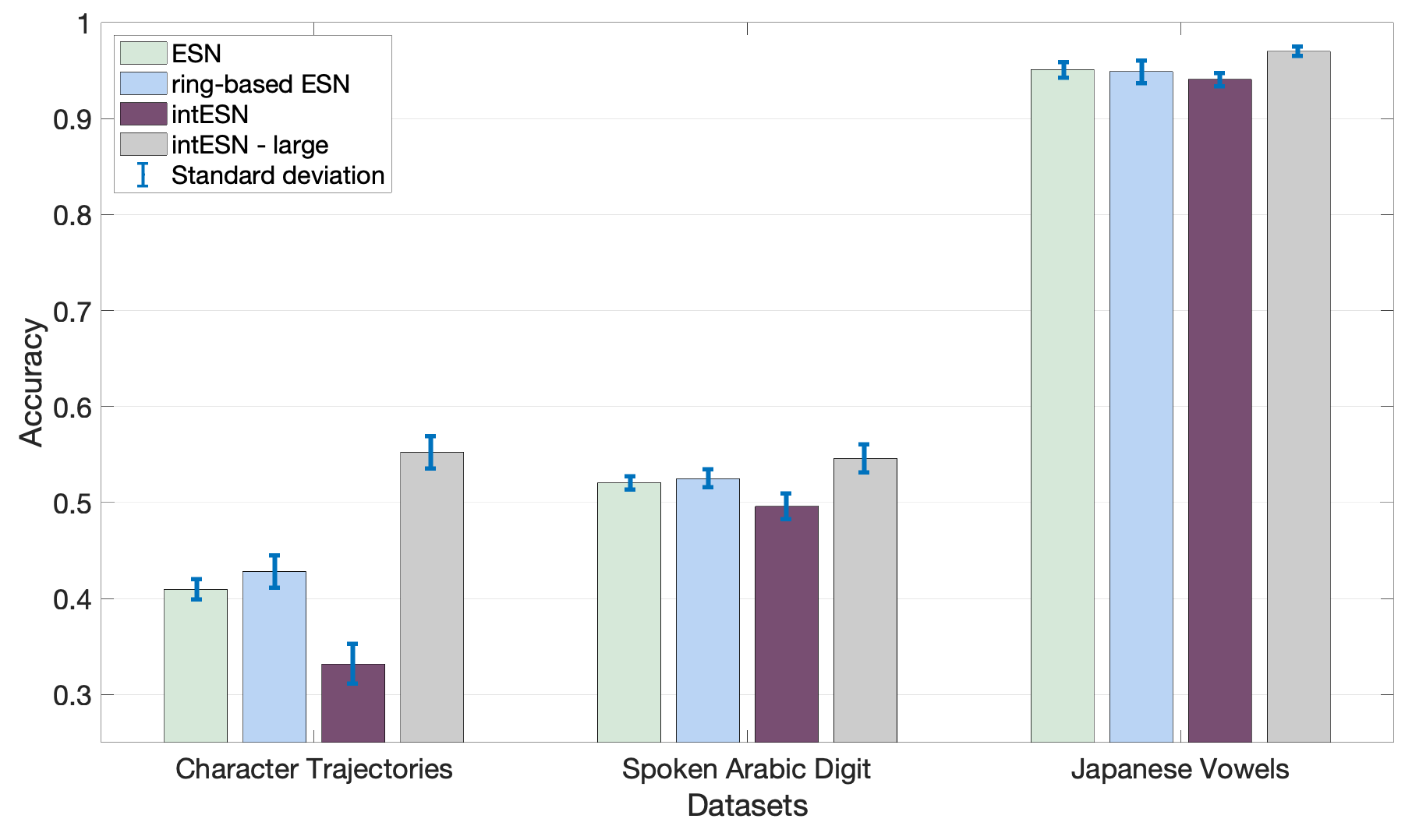}
\caption{The classification accuracy for multivariate datasets from UCI. Bars depict mean values, lines depict standard deviations. Bars denoted as ``ESN'' and ``intESN'' had the same number of neurons in their reservoirs while for ``intESN-large'' the number of neurons corresponded to ESN's memory footprint. 
}
\label{fig:multivar}
\end{figure}

%\hl{
The experimental accuracies obtained from the networks for the considered datasets are presented in Fig.~\ref{fig:univar} and ~\ref{fig:multivar}. Fig.~\ref{fig:univar} presents the results for univariate datasets while Fig.~\ref{fig:multivar} presents the results for multivariate datasets. The figures depict mean and standard deviation values across ten independent random initializations of the networks.
%{\color{brown}
Similar to subsection~\ref{sect:perf:trajectory}, the accuracy of the conventional ESN and the ring-based ESN are almost identical, thus, for the rest of this subsection we assume both of these networks when using the term ESN.  
%}

%}

%\hl{
The obtained results strongly depend on the characteristics of the data. However, it was generally observed that intESN with the memory footprint equivalent to ESN demonstrated higher classification accuracy. 
On the other hand, the classification accuracy of intESN with the same number of neurons as in ESN was similar to ESN's performance for all considered datasets but two (``Swedish Leaf'' and ``Character Trajectories'') for which the accuracy degradation was sensible.
We, therefore, conjecture that in a general case, one cannot guarantee the same classification accuracy as for
ESN. The empirical evidence, however, shows that it is not infeasible. Since placing the reported results into
the general context of time-series classifications is outside the scope of this article we do not further
elaborate on fine-tuning of hyperparameters of intESN for the best classification performance.
%On the other hand, classification accuracy of intESN with the same number of neurons as in ESN was similar to ESN's performance for most of the datasets but it %was significantly lower for two datasets: ``Swedish Leaf'' and ``Character Trajectories''. Therefore, in a general case, one cannot guarantee the same accuracy %but the empirical evidence shows that it is not unlikely. 
%Finally, it should be noted that we did not aim at placing the reported results into the general context of time-series classifications. Instead, the goal was to compare the results of intESN and ESN.
%}
%{\color{brown}
However, the interested readers are kindly referred to the Supplementary materials (Fig. S.2 and S.3) where several different values of $N$, $\kappa$, and $\rho$ were examined for each dataset. 
%}

\subsection{Modeling of dynamic processes}
\subsubsection{Learning Sinusoidal Function}

\begin{figure}[tb]%[h]%[htb]
\centering
\includegraphics[width=1.0\linewidth]{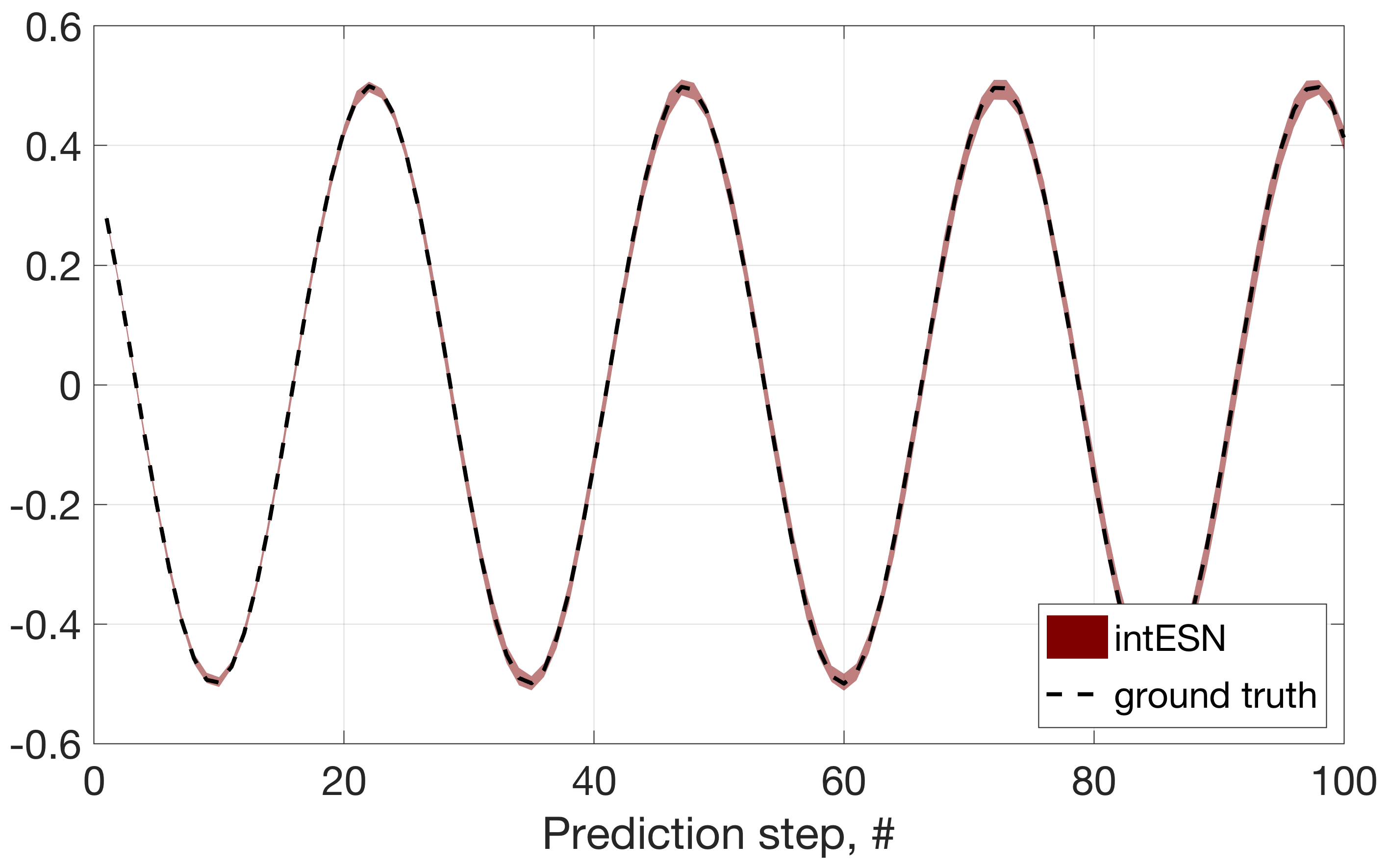}
\caption{Generation of a sinusoidal signal.}
\label{fig:Sinus}
\end{figure}

The task of learning a sinusoidal function \cite{ESN02} is
an example of a learning  simple dynamic system with the constant cyclic behavior. The 
ground truth signal was generated as follows:
~
\begin{equation}
y(n)=0.5\sin(n/4).
\label{eq:sin}
 \end{equation}
~
In this task, the input layer was not used, i.e. $K=0$ but the network
projected the activations  of the output layer back to the reservoir using $\textbf{W}^{\text{back}}$. The output
layer had only one neuron ($L=1$).  The reservoir size was fixed to $N=1000$ neurons.
The length of the training  sequence was 3000 (first 1000 steps were discarded
from the calculation). For ESN, the feedback strength for the reservoir connection matrix was set to $\rho=0.8$,
for both networks  $\lambda$ was set to 0.
A continuous value of the ground truth signal was fed-in to ESN during the training. 

For  intESN, in order to map the input signal to a bipolar vector the quantization was used. The signal was quantized as:
~
\begin{equation}
y(n)_q=\lfloor100y(n)\rceil/100.
\label{eq:quan}
 \end{equation}
~
The item memory for the projection of the output layer was populated with
bipolar vectors preserving linear  (in terms of dot product) similarity between
quantization levels \cite{Widdows15}.  The threshold for the clipping function
was set to $\kappa=3$.

\begin{figure*}[tb]%[htb]
\centering
\includegraphics[width=1.0\linewidth]{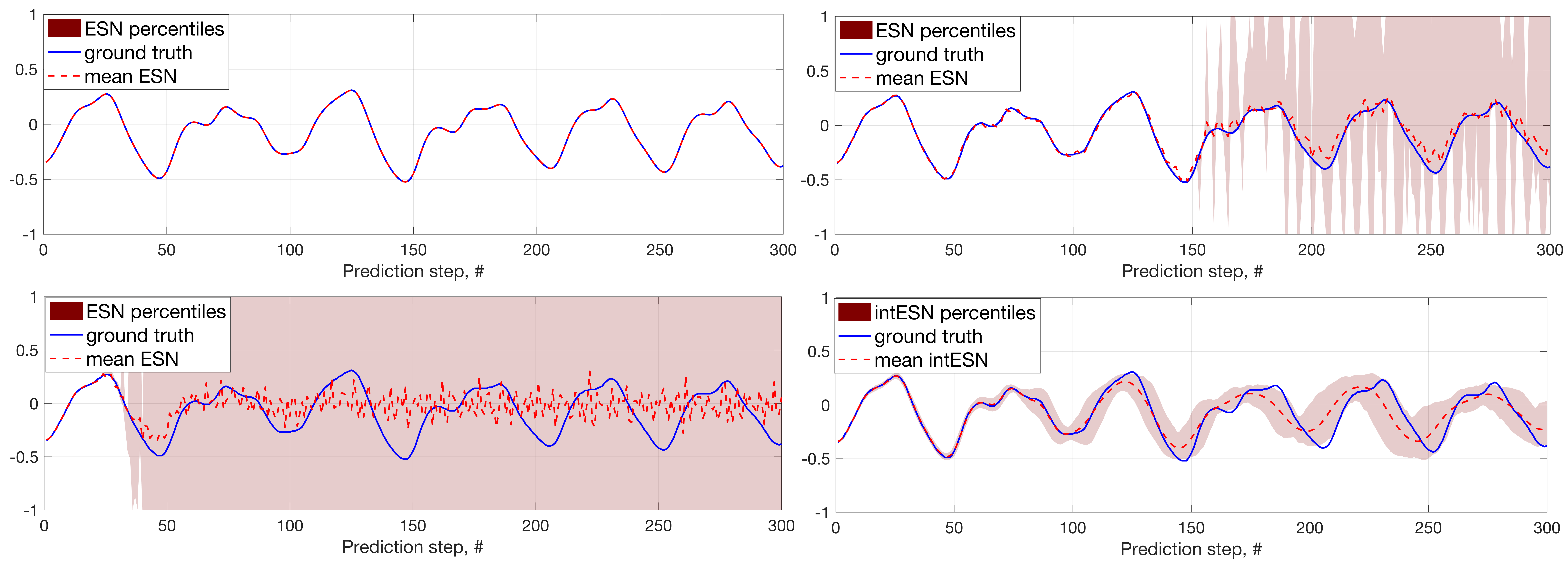}
\caption{Prediction of the Mackey-Glass series.}
\label{fig:Mackey-Glass}
\end{figure*}

In the operating phase,  the network acted as the generator of the signal
feeding its previous prediction (at time $n-1$) back to the reservoir. 
Fig.~\ref{fig:Sinus} demonstrates  the behavior of intESN during the first 100
prediction steps.  The ground truth is depicted by dashed line while the
prediction of intESN  is  illustrated by the shaded area between 10\% and 90\%
percentiles (100 simulations were performed).  The figure does not show the
performance of the conventional ESN as it just followed the ground truth without
visible deviations. intESN clearly follows the values of the ground truth
but the deviation  from the ground truth is increasing with the number of
prediction steps.  It is unavoidable for the increasing prediction horizon but,
in this scenario,  it is additionally accelerated due to the presence of the
quantization error at each prediction step. 
%{\color{brown}
It is worth noting, however, that the quality of predictions for this task could be improved by increasing the value of $\kappa=3$, i.e., at the cost of extract memory allocated for each neuron. 
The diligent readers are kindly referred to the Supplementary materials (Fig. S.4) where several different values of $\kappa$ were examined. 
%}
The next subsection will clearly demonstrate effects caused by the quantization process. 
The error is accumulated because every time when feeding the prediction back to the reservoir of intESN it should be quantized in order to fetch a vector from the item memory.

\subsubsection{Mackey-Glass series prediction}

A Mackey-Glass series is generated by the nonlinear time delay differential equation. 
It is commonly used to assess the predictive power of an RC approach.  
In this scenario, we followed  the preprocessing of data and the parameters of
ESN described in \cite{ESN04}.
The parameters of intESN  (including quantization scheme) were the same as in
the subsection above.  
%{\color{brown}
The interested readers are kindly referred to the Supplementary materials (Fig. S.5) where several different values of $N$ and $\kappa$ were examined. 
%}
The length of the training sequence was 3000 (first 1000
steps were discarded from the calculation).
Fig.~\ref{fig:Mackey-Glass} depicts  results for the first 300 prediction
steps. The results were calculated from 100 simulation runs.  The figure
includes four panels. Each panel depicts the  ground truth, the mean value of
predictions as well  as areas marking percentiles between 10\% and 90\%. The
lower right  corner corresponds to intESN while three other panels show
performance of ESN in three different cases related to the quantization of
the data.

In these scenarios ESN was trained to learn the model from the  quantized
data in order to see to which extent  it affects the network.
The upper left  corner corresponds to ESN without data quantization. In this
case, the predictions precisely follow the ground truth.  The upper right corner
corresponds to ESN trained on the quantized data but  with no quantization
during the operational phase. In such settings, the network  closely follows the
ground truth for the first 150 steps but then it often explodes.  The lower left
corner corresponds to ESN where the data was quantized during  both training
and prediction. In this scenario, the network was able produce to  produce good
prediction just for the first few dozens of steps and then entered the  chaotic
mode where even the mean value does not reflect the ground truth.  These cases
demonstrate how the quantization error could affect the predictions  especially
when it is added at each prediction step.
Note that intESN operated in  the same mode as the third ESN. Despite this
fact, its performance rather resembles that of  the second ESN where the speed
deviation of the ground truth is faster.  At the same time, the deviation of
intESN grows smoothly without a sudden explosion  in contrast to ESN.

%All examples demonstrate decent approximation of the performance achieved by the conventional ESN.

%% file: tex/FPGA.tex
%\vspace*{-0.6cm}
\section{Digital hardware experiments}
\label{sect:fpga}
%\vspace*{-0.3cm}

%{\color{red}

In order to demonstrate the amenability of intESN for digital hardware, we used a field-programmable gate array (FPGA) and implemented three different architectures: software ESN as well as hardware accelerated ESN and intESN. 
All architectures were implemented on a ZedBoard FPGA\footnote{ZedBoard. Hardware User's Guide [online], 2014. -- Available online: \url{http://zedboard.org/sites/default/files/documentations/ZedBoard_HW_UG_v2_2.pdf}.}, which contains a dual core ARM Cortex A9 CPU interfacing with a programmable logic fabric. 
The Xilinx Vivado Design Suite\footnote{Xilinx. Vivado Design Suite User Guide [online], 2018. -- Available online: \url{https://www.xilinx.com/support/documentation/sw_manuals/xilinx2018_1/ug910-vivado-getting-started.pdf}.} and Vivado SDK\footnote{Xilinx. Generating Basic Software Platforms [online], 2018. -- Available online: \url{https://www.xilinx.com/support/documentation/sw_manuals/xilinx2018_3/ug1138-generating-basic-software-platforms.pdf}.} were used to design the hardware architectures and program the FPGA board, respectively.
The efficiency evaluation (e.g., energy consumption) of the architectures was based on the recall stage of the sequence recall task  as described in Section~\ref{sect:perf:trajectory}.

\subsection{Architectures design}
\label{sect:fpga:design}

The software ESN was implemented using only CPU on the FPGA board. 
For the hardware acceleration experiments, we programed the CPU to generate the inputs and feed them into a hardware architecture (either for ESN or intESN), and then retrieve the outputs once the computation is over.
The hardware ESN and intESN architectures were designed using Vivado High Level Synthesis (HLS) using the C programming language, and later on synthesized into hardware Intellectual Property components to be integrated in a larger hardware system comprised of: the Zynq Processing System, AXI Direct Memory Access, AXI Interconnect, ESN/intESN architecture, and various other peripherals used for clocking and resetting mechanisms. 
It is important to note that for resource limitations on the ZedBoard, no pipelining or hardware optimization directives have been used on the HLS designs in order to cope with the resource usage for growing reservoir sizes and remain within the board's capacity. However, we still provide a comparison of a speed-up expected from the pipelining for the two hardware architectures.

\subsection{Evaluation methodology}

For each architecture, the reservoir size was varied between $100$, $200$, and $300$ neurons. 
In the following, we report the number of clock cycles required to accomplish the sequence recall task, the power consumption, and the area utilization (i.e., resources) required for the hardware designs.
The number of clock cycles was measured using a hardware timer incorporated into the designs. The area utilization is reported from the hardware synthesis reports provided by Vivado. 
All three architectures use the same clock frequency of $100$ MHz.
The ZedBoard contains a $10$ m$\Omega$ shunt-resistor in series with the input supply of the whole board, which could be used to obtain the overall power consumption by measuring the voltage across it. 
However, at the desired scale of changes voltage fluctuation and measurement sensitivity made it almost impossible to properly perform a precise comparison.
Therefore, instead, we used the Xilinx Power Estimator (XPE) tool\footnote{Xilinx. Power Estimator User Guide [online], 2018. -- Available online: \url{https://www.xilinx.com/support/documentation/sw_manuals/xilinx2018_3/ug440-xilinx-power-estimator.pdf}.} provided by the Vivado Design Suite to estimate the power consumption of each architecture, which is a standard option \cite{FPGASTDP2019, HACNN2018, XPEconf}.

\subsection{Efficiency evaluation results}

\begin{table}[tb]%[ht]
\renewcommand{\arraystretch}{1.3}
\caption{Area utilization of the hardware architectures.
\label{tab:fpga:resources}
\vspace{-2mm}}
    % {\scriptsize
    \begin{center}
    \begin{tabular}{|c|c|c|c|c|}\hline
        $N$ & \textbf{LUT} & \textbf{FF} & \textbf{BRAM} & \textbf{DSP48} \\ \hline \hline 
      	\multicolumn{5}{|c|}{\textbf{ESN}} \\ \hline\hline
        $100$	& $8317$	& 	$8724$ &	$31.5$ &	$47$ \\ \hline  
 	$200$ & $8368 $	& 	$8770 $ &	$87.5$ &	$47$	 \\ \hline
	$300$ 		& $15950 $	& 	$8920 $ &	$135.5$ &	$47$\\ \hline \hline   	    
        \multicolumn{5}{|c|}{\textbf{intESN}} \\ \hline\hline
	 $100$ &	$4577 $	& 	$5488 $ &	$7.5 $ &	$5 $	  \\\hline
	 $200$ 	 & $4602 $	& 	$5492 $ &	$11.5 $ &	$6 $	  \\\hline
        $300$	& $4671 $	& 	$5497 $ &	$12.5 $ &	$6 $	 \\ \hline    	
    \end{tabular}
    \end{center}
%  }
%\vspace{-5mm}
\end{table}

Table~\ref{tab:fpga:resources} presents the area utilization of the hardware architectures for different sizes of reservoir.
It is clear that the resource utilization of the hardware ESN is always larger than that of hardware intESN. 
This is an empirical manifestation of the facts that intESN: a) requires a lower memory footprint ($\kappa=3$) and b) its machinery uses simpler operations, e.g., the clipping instead of $\tanh()$ and the cyclic shift instead of vector-matrix multiplication. 
It is important to note, that the drastic increase of LUT utilization when the reservoir size of ESN was set to $300$ is due to resource constraints on the FPGA board since the total number of usable BRAM units is 140.
Thus, the increased number of LUTs was used as an alternative way of increasing the memory capacity of the board.

\begin{table}[tb]%[ht]
\renewcommand{\arraystretch}{1.3}
\caption{Number of clock cycles (time) involved in the sequence recall task.
\label{tab:fpga:speed}
\vspace{-2mm}}
    % {\scriptsize
    \begin{center}
    \begin{tabular}{|c|c|c|c|}\hline
        $N$ & \textbf{software ESN} & \textbf{hardware ESN} & \textbf{hardware intESN}  \\ \hline 
        $100$	&  \num{3.0e8}  (3.04 \si{s}) 	& 	\num{5.8e8}  (5.77 \si{s})  &	 \num{1.4e8}  (1.42 \si{s})   \\ \hline  
 	$200$ & \num{1.0e9}  (9.99 \si{s}) 	& 	\num{1.9e9}  (18.74 \si{s}) &	\num{2.8e8}  (2.82 \si{s})  \\ \hline
	$300$ & \num{2.1e9}  (21.16 \si{s}) 	& 	\num{3.9e9}  (38.90 \si{s}) &\num{4.5e8}  (4.46 \si{s})  \\ \hline 	    

    \end{tabular}
    \end{center}
%  }
%\vspace{-5mm}
\end{table}

Table~\ref{tab:fpga:resources} presents the number of clock cycles (time in seconds) necessary to perform the operating phase of the sequence recall task for each architecture.
The number of clock cycles was measured with the help of the hardware timer.
The time was calculated using the known frequency of the clock ($100$ MHz)
As expected, for each architecture the operation time increases with the increased reservoir size. 
However, for any reservoir size, intESN is several times faster than both implementations of ESN (at least $2.1$ against the software ESN when $N=100$). The gain is increasing with the increased reservoir size so that when $N=300$ the hardware intESN is $8.7$ times faster compared to the hardware ESN. 

An interesting remark in Table~\ref{tab:fpga:speed} is that the hardware ESN seems slower than the software ESN. 
It is counterintuitive since the hardware architecture is expected to accelerate the computations. 
In the consider case, this fact is explained by the absence of pipelining (cf. Section~\ref{sect:fpga:design}) what prevents further hardware optimizations.
The pipelining would certainly increase the speed at the price of the increased area utilization, which would make the hardware designs being inconceivable on the target board.
In order to evaluate the effect of the pipelining on the speed of the hardware architectures, we have used the Vivado HLS synthesis report to estimate the number of clock cycles per iteration when $N=300$. 
The results are presented in Table~\ref{tab:fpga: pipelining}. 
It shows that the pipelining significantly decreases the number of clock cycles, which makes the hardware ESN much faster than the software one. 
At the same time, the speed-up obtained for the hardware intESN is still higher than that of the hardware ESN, which makes intESN even more efficient than ESN.

\begin{figure}[tb]%[!ht]%[t!]
\centering
\includegraphics[width=1.0\linewidth]{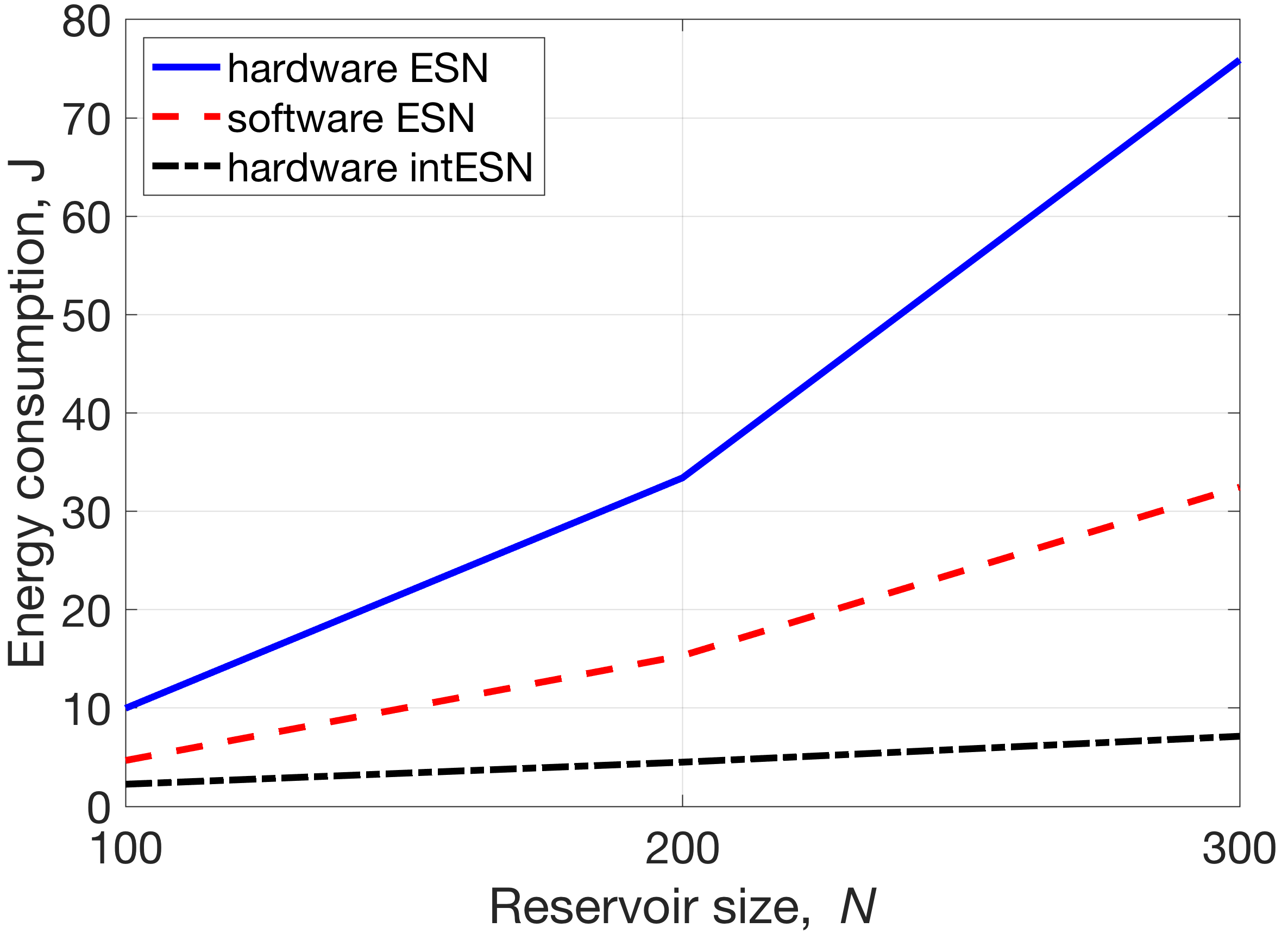}
\caption{
The overall energy consumption of all three architectures against the reservoir size. 
}
\label{fig:fpga:power}
\end{figure}

\begin{table}[tb]%[ht]
\renewcommand{\arraystretch}{1.3}
\caption{Speed-up comparison with the use of pipelining.
\label{tab:fpga: pipelining}
\vspace{-2mm}}
    % {\scriptsize
    \begin{center}
    \begin{tabular}{|c|c|c|c|c|}\hline
	 & \multicolumn{2}{c|}{\textbf{ESN}} & \multicolumn{2}{c|}{\textbf{intESN}} \\ \hline
        Pipelining & \cmark & \xmark & \cmark & \xmark\\ \hline       	 
       Clock cycles	& $1,297,072$	& 	$51,948$ &	$148,241$ &	$2,856$ \\ \hline  
	Speed-up & \multicolumn{2}{c|}{$24.97$} & \multicolumn{2}{c|}{$51.91$} \\ \hline
    \end{tabular}
    \end{center}
%  }
%\vspace{-5mm}
\end{table}

\begin{table}[tb]%[ht]
\renewcommand{\arraystretch}{1.3}
\caption{XPE overall power consumption (Watts)
\label{tab:fpga:power}
\vspace{-2mm}}
    % {\scriptsize
    \begin{center}
    \begin{tabular}{|c|c|c|c|}\hline
        $N$ & \textbf{software ESN} & \textbf{hardware ESN} & \textbf{hardware intESN}  \\ \hline 
        $100$	& $1.53$ 	& 	$1.73$  &	 $1.59$    \\ \hline  
 	$200$ & $1.53$ 	& 	$1.78$ &	$1.60$  \\ \hline
	$300$ & $1.53$ 	& 	$1.95$ & $1.60$ \\ \hline 	    

    \end{tabular}
    \end{center}
%  }
%\vspace{-5mm}
\end{table}

Table~\ref{tab:fpga:power} presents  the power consumption of each architecture. 
The first observation is that the consumption of the  software ESN remains constant for different $N$.
This is because the hardware itself remains fixed, and the software computations that are performed on the Cortex CPU are the same for each configuration. 
Moreover, the software ESN has the lowest power consumption because it is only comprised of the CPU and the hardware timer, whereas the other two include other hardware components in addition to those. 
However, for the hardware implementation of ESN one could see a noticeable increase for a larger reservoir sizes. There is also an increase for the hardware intESN but it is slower compared to the hardware ESN. 
Moreover, for the fixed reservoir size, intESN's power consumption remains lower than that of ESN.

Fig.~\ref{fig:fpga:power} depicts the overall energy consumption of each architecture for different sizes of reservoir.
The energy consumption was calculated as a product of the overall power consumption reported in Table~\ref{tab:fpga:power} and the operating time reported in Table~\ref{tab:fpga:speed}. 
Fig.~\ref{fig:fpga:power} clearly demonstrates that the energy efficiency of the hardware intESN is much better than that of both the hardware and software ESNs.
Scaling the reservoir size leads to an increase of the energy consumption of all architectures.
However, the slope of the curve for intESN is lower than for ESN's architectures. 
Therefore, the energy saved by the use of intESN drastically increases with increasing size of reservoir (e.g., for $N=300$ it needs $10.6$ times lower energy than for the hardware ESN), which is a strong argument in favor of intESN. 
%}

%\todo[inline, color=olive]{
%DK: Redraw Fig.~\ref{fig:fpga:power} to get better quality \\
%}

%% file: tex/Discussion.tex
\section{Discussion}
\label{sect:dis}

\subsubsection{Efficiency of intESN}
%{\color{red}
In principle, evaluation of the computational efficiency of the proposed intESN could be done even in a simplistic manner using only a computer.
For example, we have performed the initial assessment by simplified execution time measurements with Matlab.
We used our Matlab implementations of both networks for the trajectory association task (Section~\ref{sect:perf:trajectory}) with $N=300$ neurons to compare the times of projecting data and executing a reservoir. ESN was implemented using 32-bit float type (type \textit{single} in Matlab) while intESN was implemented using 8-bits integer type (type \textit{int8} in Matlab).
On average, the time required by intESN was $3.9$ times less than that of ESN.
Thus, even the training time needed for convergence with intESN is shorter than that of ESN because intESN is much faster when it comes to projecting data and executing a reservoir while the time for estimating the readout matrix would be the same for both networks. 
%}

%{\color{red}
Section~\ref{sect:fpga} presented the proper evaluation of the computational efficiency of the proposed intESN approach and the conventional ESN using the FPGA board. The hardware implementation of intESN was compared to two reference implementations: software ESN and hardware ESN. 
The detailed benchmarking tests have supported our claims about the efficiency of intESN. 
%}
%{\color{brown}
It is also worth noting that the efficiency gains of the intESN would be less relative to the ring-based ESN as essentially both networks use the same efficient mechanism for reservoir's connectivity. 
%}

%\hl{
%It is hard to make claims without a proper implementation design and benchmarking tests. However, for the sake %of discussion, we conjecture that given the implementation intESN could be more efficient than ESN. 

\subsubsection{Hyperparameters}

%\hl{
%{\color{red}
In comparison to ESN, intESN has fewer hyperparameters. The common hyperparameter is the reservoir size $N$. Two specific intESN hyperparameters are clipping threshold $\kappa$ and the mapping to the reservoir; while for ESN we have to choose $\rho$, $\beta$, and $\alpha$.
When $N$ is fixed then $\kappa$ in intESN has an effect on the network's memory similar to $\rho$ and $\beta$ in ESN (see the next subsection for details). 
However, the difference is that $\kappa$ takes only positive integer values. 
%}
This has its pros and cons. It could be much easier to optimize a single hyperparameter in the integer range.
On the other hand, having real-valued hyperparameters one can get a configuration providing much finer tuning of network's memory for a given task.  

With respect to the mapping (projection) to the reservoir, it is probably the most non-trivial part on intESN. Especially, when data to be projected are real numbers. In this article, we mention three different strategy for mapping real numbers to bipolar or ternary vectors: puncturing of non-zero elements (Section~\ref{sect:perf:analog}), mapping preserving linear similarity between vectors \cite{Widdows15}, and nonlinear approximate mapping using scatter codes \cite{TNNLS18, scatter}. We suggest that it is a useful heuristic to try each of the approaches and choose the one performing best.
%}

\subsubsection{On equivalence of ESN and intESN in terms of forgetting time constants}

%\hl{In fact, this work can also be conceived as an engineering design following from the recent theoretical results in~\cite{Frady17} where rigorous connections between RC and HDC were presented.
%}

\begin{figure}[tb]%[!ht]%[t!]
\centering
\includegraphics[width=0.8\columnwidth]{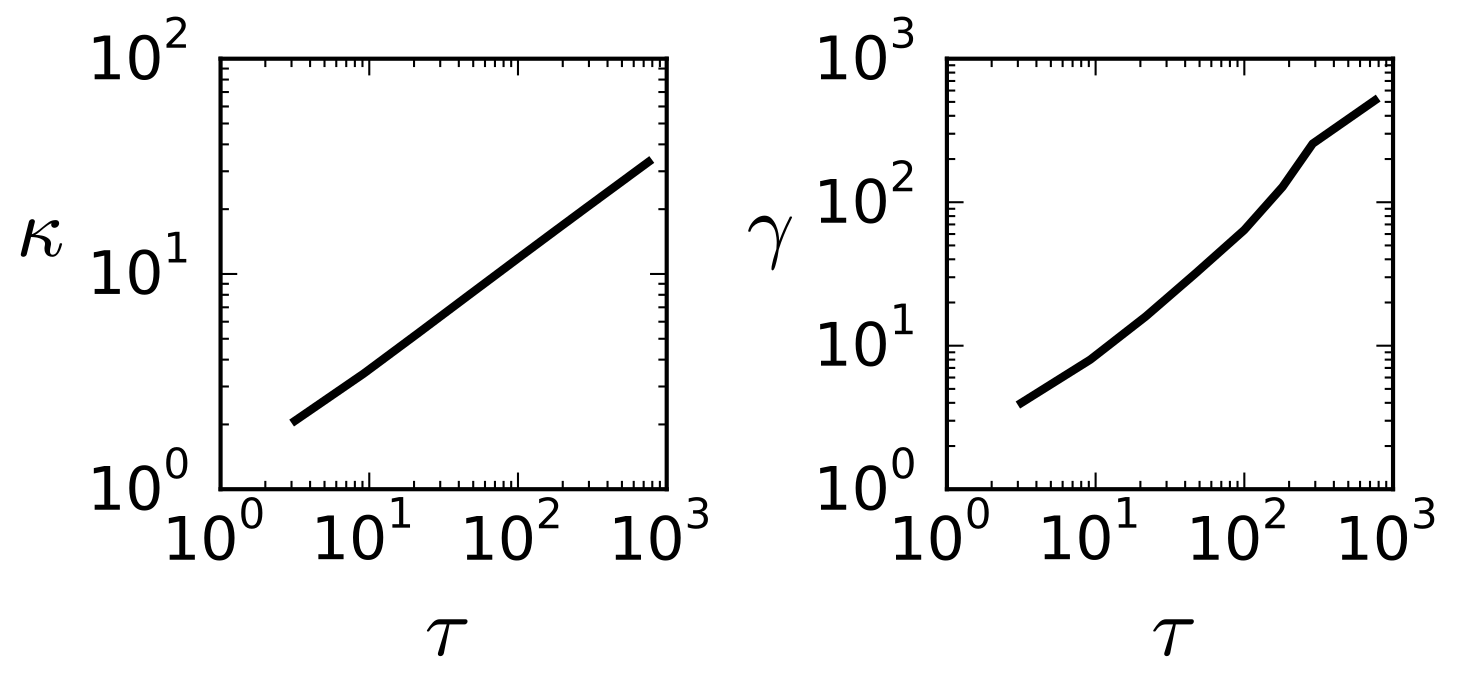}
\caption{
%\hl{
Correspondence between time-constants of intESN and special cases of ESN.
%}
}
\label{fig:time:const}
%\vspace*{-0.5cm}
\end{figure}

%\hl{
Section~\ref{sect:perf:trajectory} presented the experimental comparison of storage capabilities of intESN and ESN. 
An analytical approach to the treatment of memory capacity of reservoir was presented in~\cite{Frady17} (Section~2.4.4). 
%In fact, it is also possible to perform a more analytical analysis as it is was shown 
The work introduced an analytical tool called the forgetting time constant (denoted as $\tau$). The forgetting time constant is a scalar characterizing the memory decay in a network. In the case of intESN, it can be calculated analytically using the clipping threshold $\kappa$. For ESN, currently only special cases can be analyzed. For example, when reservoir neurons are linear the feedback strength $\rho$ will determine $\tau$. The other example is when reservoir neurons use $\tanh$, the feedback strength is fixed to one, and the reservoir update rule is modified to $f(\textbf{x})=\gamma \tanh( \textbf{x} / \gamma)$, where $\gamma$ denotes a gain parameter. This parameter affects $\tau$, which could be calculated numerically. Fig.~\ref{fig:time:const} presents the implicit comparison of this special case of ESN and intESN via their forgetting time constants which are determined by $\gamma$ and $\kappa$ respectively. Both parameters similarly (not far from linear in logarithmic coordinates) affect $\tau$. It allows arguing that the networks are close to being functionally identical in terms of storage capabilities. The development of an approach for estimating the forgetting time constant for ESN in a general case is a part of our agenda for the future work.
%}

\subsubsection{Training the readout in a generator mode}

%\hl{
In the experiments generating time-series, we used the teacher forcing approach for training a readout matrix. This, however, does not have to be the compulsory choice for intESN. We do not foresee any complications for applying other approaches allowing to modify the network's behavior for producing complex target functions. In particular, FORCE method \cite{Sussillo2009}, which uses error-based modification of readout weights during the training process, can be used as it has a mode in which only weights of a readout matrix are changed while leaving the rest of the network fixed. 
%}

%% file: tex/Conclusions.tex
\section{Conclusions}
\label{sect:conclusions}

In this article we  proposed an architecture for integer approximation of the
reservoir computing, which is based on the  mathematics of  hyperdimensional
computing. The neurons in the reservoir are described by integers in the limited range and the
update operations include only addition, permutation (cyclic shift), and clipping.  Therefore, the integer Echo State Network
has substantially smaller memory footprint and higher computational efficiency
compared to the conventional Echo State Network with the same number of neurons in
the reservoir. The actual number of bits for representing a neuron depends on
the clipping threshold $\kappa$, but can be significantly lower than 32-bit floats in Echo State Network. For example, in our experiments the results were
obtained with $\kappa=3$ and $\kappa=7$, which effectively makes it sufficient to represent a
neuron with only three or four bits respectively. We demonstrated that the performance of the integer Echo State Network is
comparable to the conventional Echo State Network in terms of memory capacity, potential capabilities for classification of time-series and modeling dynamic systems. 
The better performance was observed when the memory footprint of reservoir of the integer Echo State Network was set to that of the conventional Echo State Network.
%{\color{red}
The experiment on the digital hardware have validated the amenability of the integer Echo State Network for significantly improving the energy efficiency of computations. 
%}
Further improvements can be made by optimization of the parameters and better quantization schemes for handling continuous values.
 Naturally, due to the
peculiarity of input data projection into the integer Echo State Network, the performance of the
network 
in tasks for modeling dynamic systems
is to a certain degree lower than that of the  conventional Echo State Network. This,
however, does not undermine the importance of integer Echo State Networks,  which are  extremely
attractive for memory and power savings, and in the general
area of approximate computing,  where errors and  approximations are becoming
acceptable as  long as the outcomes have a well-defined statistical behavior. 
%\cite{HDNP17}.

%\textbf{Acknowledgements.} This study  is  supported in part  by the Swedish
%Research Council (grant no. 2015-04677).
%The authors thank Ozgur Yilmaz for fruitful discussions during BICA2016 on the
%usage of cellular automata in the scope of hyperdimensional computing, which
%inspired the current work and Niklas Karvonen  for general discussions on
%cellular automata.

%% file: main.bbl
% Generated by IEEEtran.bst, version: 1.14 (2015/08/26)
\begin{thebibliography}{10}
\providecommand{\url}[1]{#1}
\csname url@samestyle\endcsname
\providecommand{\newblock}{\relax}
\providecommand{\bibinfo}[2]{#2}
\providecommand{\BIBentrySTDinterwordspacing}{\spaceskip=0pt\relax}
\providecommand{\BIBentryALTinterwordstretchfactor}{4}
\providecommand{\BIBentryALTinterwordspacing}{\spaceskip=\fontdimen2\font plus
\BIBentryALTinterwordstretchfactor\fontdimen3\font minus
  \fontdimen4\font\relax}
\providecommand{\BIBforeignlanguage}[2]{{%
\expandafter\ifx\csname l@#1\endcsname\relax
\typeout{** WARNING: IEEEtran.bst: No hyphenation pattern has been}%
\typeout{** loaded for the language `#1'. Using the pattern for}%
\typeout{** the default language instead.}%
\else
\language=\csname l@#1\endcsname
\fi
#2}}
\providecommand{\BIBdecl}{\relax}
\BIBdecl

\bibitem{RC09}
{M. Lukosevicius and H. Jaeger}, ``{Reservoir Computing Approaches to Recurrent
  Neural Network Training},'' \emph{Computer Science Review}, vol.~{3},
  no.~{3}, pp. {127--149}, {2009}.

\bibitem{Sussillo2009}
{D. Sussillo and L.F. Abbott}, ``{Generating Coherent Patterns of Activity from
  Chaotic Neural Networks},'' \emph{Neuron}, vol.~63, no.~4, pp. 544--557,
  2009.

\bibitem{ESN11NIPS}
F.~Triefenbach, A.~Jalalvand, B.~Schrauwen, and J.-P. Martens, ``{Phoneme
  Recognition with Large Hierarchical Reservoirs},'' in \emph{{Advances in
  Neural Information Processing Systems (NIPS) 23}}, 2010, pp. 2307--2315.

\bibitem{ESN04}
{H. Jaeger and H. Haas}, ``{Harnessing Nonlinearity: Predicting Chaotic Systems
  and Saving Energy in Wireless Communication},'' \emph{Science}, vol. {304},
  no. {5667}, pp. {78--80}, {2004}.

\bibitem{RCnature11}
{L. Appeltant and M.C. Soriano and G. Van der Sande and J. Danckaert and S.
  Massar and J. Dambre and B. Schrauwen and C.R. Mirasso and I. Fischer},
  ``{Information Processing Using a Single Dynamical Node as Complex System},''
  \emph{Nature Communications}, vol.~{2}, no.~{1}, pp. {1--6}, {2011}.

\bibitem{CHAOS18}
J.~Pathak, B.~Hunt, M.~Girvan, Z.~Lu, and E.~Ott, ``{Model-Free Prediction of
  Large Spatiotemporally Chaotic Systems from Data: A Reservoir Computing
  Approach},'' \emph{Physical Review Letters}, vol. 120, pp. 1--5, 2018.

\bibitem{ECCV16}
M.~Rastegari, V.~Ordonez, J.~Redmon, and A.~Farhadi, ``{XNOR-Net: ImageNet
  Classification Using Binary Convolutional Neural Networks},'' in
  \emph{{European Conference on Computer Vision}}, ser. Lecture Notes in
  Computer Science, vol. 9908, 2016, pp. 525--542.

\bibitem{BinNN}
I.~Hubara, M.~Courbariaux, D.~Soudry, R.~El-Yaniv, and Y.~Bengio, ``{Binarized
  Neural Networks},'' in \emph{{Advances in Neural Information Processing
  Systems (NIPS) 29}}, 2016, pp. 1--9.

\bibitem{QuanNN}
------, ``Quantized neural networks: Training neural networks with low
  precision weights and activations,'' \emph{Journal of Machine Learning
  Research}, pp. 1--30, 2018.

\bibitem{Frady17}
E.~P. Frady, D.~Kleyko, and F.~T. Sommer, ``{A Theory of Sequence Indexing and
  Working Memory in Recurrent Neural Networks},'' \emph{Neural Computation},
  vol.~30, pp. 1449--1513, 2018.

\bibitem{Kanerva09}
P.~Kanerva, ``{Hyperdimensional Computing: An Introduction to Computing in
  Distributed Representation with High-Dimensional Random Vectors},''
  \emph{Cognitive Computation}, vol.~1, no.~2, pp. 139--159, 2009.

\bibitem{Gayler2003}
R.~Gayler, ``{Vector Symbolic Architectures Answer Jackendoff's Challenges for
  Cognitive Neuroscience},'' in \emph{{Proceedings of the Joint International
  Conference on Cognitive Science. ICCS/ASCS}}, {}, Ed.\hskip 1em plus 0.5em
  minus 0.4em\relax {}, 2003, pp. 133--138.

\bibitem{Bengio94}
Y.~Bengio, P.~Simard, and P.~Frasconi, ``{Learning Long-Term Dependencies with
  Gradient Descent is Difficult},'' \emph{IEEE Transactions on Neural
  Networks}, vol.~5, no.~2, pp. 157--166, 1994.

\bibitem{LSTM97}
{S. Hochreiter and J. Schmidhuber}, ``{Long Short-Term Memory},'' \emph{Neural
  Computation}, vol.~{9}, no.~{8}, pp. {1735--1780}, {1997}.

\bibitem{LSM02}
{W. Maass and T. Natschlager and H. Markram}, ``{Real-Time Computing Without
  Stable States: A New Framework for Neural Computation Based on
  Perturbations},'' \emph{Neural Computation}, vol.~{14}, no.~{11}, pp.
  {2531--2560}, {2002}.

\bibitem{ESN03}
H.~Jaeger, ``{Adaptive Nonlinear System Identification with Echo State
  Networks},'' in \emph{{Advances in Neural Information Processing Systems
  (NIPS) 15}}, 2003, pp. 593--600.

\bibitem{ELM06}
{G. Huang and Q. Zhu and C. Siew}, ``{Extreme Learning Machine: Theory and
  Applications},'' \emph{{Neurocomputing}}, vol.~{70}, no. {1-3}, pp.
  {489--501}, {2006}.

\bibitem{ELM15}
{G. Huang}, ``{What are Extreme Learning Machines? Filling the Gap Between
  Frank Rosenblatt's Dream and John von Neumann's Puzzle},'' \emph{{Cognitive
  Computation}}, vol.~7, pp. 263--278, 2015.

\bibitem{Jaeger2001}
H.~Jaeger, ``{The ``Echo State'' Approach to Analysing and Training Recurrent
  Neural Networks},'' GMD Report 148, German National Research Center for
  Information Technology, Tech. Rep., 2001.

\bibitem{Weddell2008}
S.~J. Weddell and R.~Y. Webb, ``{Reservoir Computing for Prediction of the
  Spatially-Variant Point Spread Function},'' \emph{IEEE Journal of Selected
  Topics in Signal Processing}, vol.~2, no.~5, pp. 624--634, 2008.

\bibitem{Antonik2017}
P.~Antonik, M.~Haelterman, and S.~Massar, ``{Brain-Inspired Photonic Signal
  Processor for Generating Periodic Patterns and Emulating Chaotic Systems},''
  \emph{Physical Review Applied}, vol.~7, no.~5, pp. 1--16, 2017.

\bibitem{ESNEEG}
E.~M. Forney, C.~W. Anderson, W.~J. Gavin, P.~L. Davies, M.~C. Roll, and B.~K.
  Taylor, ``{Echo State Networks for Modeling and Classification of EEG Signals
  in Mental-Task Brain-Computer Interfaces},'' Colorado State University,
  CS-15-102, Tech. Rep., November 2015.

\bibitem{ComparisonReadOut}
A.~Prater, ``{Comparison of Echo State Network Output Layer Classification
  Methods on Noisy Data},'' in \emph{International Joint Conference on Neural
  Networks (IJCNN)}, May 2017, pp. 2644--2651.

\bibitem{Bianchi2018}
F.~M. Bianchi, S.~Scardapane, S.~Lokse, and R.~Jenssen, ``{Reservoir Computing
  Approaches for Representation and Classification of Multivariate Time
  Series},'' \emph{IEEE Transactions on Neural Networks and Learning Systems},
  vol.~PP, no.~99, pp. 1--11, 2020.

\bibitem{Yilmaz15a}
O.~Yilmaz, ``{Machine Learning Using Cellular Automata Based Feature Expansion
  and Reservoir Computing},'' \emph{{Journal of Cellular Automata}}, vol.~10,
  no. 5-6, pp. 435--472, 2015.

\bibitem{ISBI}
D.~Kleyko, S.~Khan, E.~Osipov, and S.~P. Yong, ``{Modality Classification of
  Medical Images with Distributed Representations based on Cellular Automata
  Reservoir Computing},'' in \emph{{IEEE International Symposium on Biomedical
  Imaging}}, 2017, pp. 1--4.

\bibitem{Yilmaz15b}
O.~Yilmaz, ``{Symbolic Computation Using Cellular Automata-Based
  Hyperdimensional Computing},'' \emph{{Neural Computation}}, vol.~27, no.~12,
  pp. 2661--2692, 2015.

\bibitem{CAHD17}
S.~Nichele and A.~Molund, ``{Deep Learning with Cellular Automaton-Based
  Reservoir Computing},'' \emph{Complex Systems}, vol.~26, no.~4, pp. 319--340,
  2017.

\bibitem{RCELMCA17}
N.~McDonald, ``{Reservoir Computing and Extreme Learning Machines using Pairs
  of Cellular Automata Rules},'' in \emph{International Joint Conference on
  Neural Networks (IJCNN)}, 2017, pp. 2429--2436.

\bibitem{Han2018}
M.~Han and M.~Xu, ``{Laplacian Echo State Network for Multivariate Time Series
  Prediction},'' \emph{IEEE Transactions on Neural Networks and Learning
  Systems}, vol.~29, no.~1, pp. 238--244, 2018.

\bibitem{Qiao2017}
J.~Qiao, F.~Li, H.~Han, and W.~Li, ``{Growing Echo-State Network with Multiple
  Subreservoirs},'' \emph{IEEE Transactions on Neural Networks and Learning
  Systems}, vol.~28, no.~2, pp. 391--404, 2017.

\bibitem{BianchiTNNLS18}
F.~Bianchi, L.~Livi, and C.~Alippi, ``{Investigating Echo-State Networks
  Dynamics by Means of Recurrence Analysis},'' \emph{IEEE Transactions on
  Neural Networks and Learning Systems}, vol.~29, no.~2, pp. 427--439, 2018.

\bibitem{Livi2018}
L.~Livi, F.~Bianchi, and C.~Alippi, ``{Determination of the Edge of Criticality
  in Echo State Networks Through Fisher Information Maximization},'' \emph{IEEE
  Transactions on Neural Networks and Learning Systems}, vol.~29, no.~3, pp.
  706--717, 2018.

\bibitem{Ozturk2007}
M.~Ozturk, D.~Xu, and J.~Principe, ``{Analysis and Design of Echo State
  Networks},'' \emph{Neural Computation}, vol.~19, pp. 111--138, 2007.

\bibitem{Busing2010}
L.~Busing, B.~Schrauwen, and R.~Legenstein, ``{Connectivity, Dynamics, and
  Memory in Reservoir Computing with Binary and Analog Neurons},'' \emph{Neural
  Computation}, vol.~22, pp. 1272--1311, 2010.

\bibitem{Strauss2012}
T.~Strauss, W.~Wustlich, and R.~Labahn, ``{Design Strategies for Weight
  Matrices of Echo State Networks},'' \emph{Neural Computation}, vol.~24,
  no.~12, pp. 3246--3276, 2012.

\bibitem{Tino00}
{S. Parfitt and P. Ti\v{n}o and G. Dorffner}, ``{Graded Grammaticality in
  Prediction Fractal Machines},'' in \emph{{Advances in Neural Information
  Processing Systems (NIPS) 14}}, 2000, pp. 52--58.

\bibitem{Tino04}
P.~Ti\v{n}o, M.~Cerna\v{n}sky, and L.~Be\v{n}uskova, ``{Markovian Architectural
  Bias of Recurrent Neural Networks},'' \emph{IEEE Transactions on Neural
  Networks}, vol.~15, no.~1, pp. 6--15, 2004.

\bibitem{Tino07}
{P. Ti\v{n}o and B. Hammer and M. Boden}, ``{Markovian Bias of Neural-Based
  Architectures with Feedback Connections},'' in \emph{{Perspectives of
  Neural-Symbolic Integration}}, ser. Studies in Computational Intelligence,
  vol.~77, 2007, pp. 95--133.

\bibitem{MinESN}
A.~Rodan and P.~Ti\v{n}o, ``{Minimum Complexity Echo State Network},''
  \emph{IEEE Transactions on Neural Networks}, vol.~22, no.~1, pp. 131--144,
  2011.

\bibitem{ESNAnomaly09}
{M. Chang and A. Terzis and P. Bonnet}, ``{Mote-Based Online Anomaly Detection
  Using Echo State Networks},'' in \emph{{Distributed Computing in Sensor
  Systems (DCOSS)}}, ser. Lecture Notes in Computer Science, vol. 5516, 2009,
  pp. 72--86.

\bibitem{Bacciu2013}
{D. Bacciu and S. Chessa and C. Gallicchio and A. Micheli and P. Barsocchi},
  ``{An Experimental Evaluation of Reservoir Computation for Ambient Assisted
  Living},'' in \emph{{Neural Nets and Surroundings: 22nd Italian Workshop on
  Neural Nets}}, ser. Smart Innovation, Systems and Technologies, vol.~19,
  2013, pp. 41--50.

\bibitem{Bacciu2014}
D.~Bacciu, P.~Barsocchi, S.~Chessa, C.~Gallicchio, and A.~Micheli, ``{An
  Experimental Characterization of Reservoir Computing in Ambient Assisted
  Living Applications},'' \emph{Neural Computing and Applications}, vol.~24,
  pp. 1451--1464, 2014.

\bibitem{ESNtut12}
M.~Lukosevicius, ``{A Practical Guide to Applying Echo State Networks},'' in
  \emph{{Neural Networks: Tricks of the Trade}}, ser. Lecture Notes in Computer
  Science, vol. 7700, 2012, pp. 659--686.

\bibitem{PlateTr}
T.~A. Plate, ``{Holographic Reduced Representations},'' \emph{IEEE Transactions
  on Neural Networks}, vol.~6, no.~3, pp. 623--641, 1995.

\bibitem{Gallant}
S.~I. Gallant and T.~W. Okaywe, ``{Representing Objects, Relations, and
  Sequences},'' \emph{Neural Computation}, vol.~25, no.~8, pp. 2038--2078,
  2013.

\bibitem{MAP}
R.~W. Gayler, ``{Multiplicative Binding, Representation Operators \&
  Analogy},'' in \emph{Gentner, D., Holyoak, K. J., Kokinov, B. N. (Eds.),
  Advances in analogy research: Integration of theory and data from the
  cognitive, computational, and neural sciences}, New Bulgarian University,
  Sofia, Bulgaria, 1998, pp. 1--4.

\bibitem{Gallant2016}
S.~I. Gallant and P.~Culliton, ``{Positional Binding with Distributed
  Representations},'' in \emph{{International Conference on Image, Vision and
  Computing (ICIVC)}}, 2016, pp. 108--113.

\bibitem{PlateBook}
T.~A. Plate, \emph{{Holographic Reduced Representations: Distributed
  Representation for Cognitive Structures}}.\hskip 1em plus 0.5em minus
  0.4em\relax {Stanford: Center for the Study of Language and Information
  (CSLI)}, 2003.

\bibitem{Rachkovskij2001}
D.~A. Rachkovskij, ``{Representation and Processing of Structures with Binary
  Sparse Distributed Codes},'' \emph{IEEE Transactions on Knowledge and Data
  Engineering}, vol.~3, no.~2, pp. 261--276, 2001.

\bibitem{HD_ICRC16}
A.~Rahimi, S.~Benatti, P.~Kanerva, L.~Benini, and J.~M. Rabaey,
  ``{Hyperdimensional Biosignal Processing: A Case Study for EMG-based Hand
  Gesture Recognition},'' in \emph{2016 IEEE International Conference on
  Rebooting Computing (ICRC)}, Oct 2016, pp. 1--8.

\bibitem{KanervaBook}
P.~Kanerva, \emph{{Sparse Distributed Memory}}.\hskip 1em plus 0.5em minus
  0.4em\relax The MIT Press, 1988.

\bibitem{Kleyko2015}
D.~Kleyko, E.~Osipov, A.~Senior, A.~I. Khan, and Y.~A. Sekercioglu,
  ``{Holographic Graph Neuron: A Bio-inspired Architecture for Pattern
  Processing},'' \emph{IEEE Transactions on Neural Networks and Learning
  Systems}, vol.~28, no.~6, pp. 1250--1262, 2017.

\bibitem{Scalarencoding}
D.~A. Rachkovskij, S.~V. Slipchenko, E.~M. Kussul, and T.~N. Baidyk, ``{Sparse
  Binary Distributed Encoding of Scalars},'' \emph{Journal of Automation and
  Information Sciences}, vol.~37, no.~6, pp. 12--23, 2005.

\bibitem{Widdows15}
D.~Widdows and T.~Cohen, ``{Reasoning with Vectors: A Continuous Model for Fast
  Robust Inference},'' \emph{Logic Journal of the IGPL}, vol.~23, no.~2, pp.
  141--173, 2015.

\bibitem{TNNLS18}
D.~Kleyko, A.~Rahimi, D.~Rachkovskij, E.~Osipov, and J.~Rabaey,
  ``{Classification and Recall with Binary Hyperdimensional Computing:
  Tradeoffs in Choice of Density and Mapping Characteristics},'' \emph{IEEE
  Transactions on Neural Networks and Learning Systems}, vol.~29, no.~12, pp.
  5880--5898, 2018.

\bibitem{intSOM}
D.~{Kleyko}, E.~{Osipov}, D.~{De Silva}, U.~{Wiklund}, and D.~{Alahakoon},
  ``{Integer Self-Organizing Maps for Digital Hardware},'' in
  \emph{International Joint Conference on Neural Networks (IJCNN)}, 2019, pp.
  1--8.

\bibitem{intRVFL}
D.~Kleyko, M.~Kheffache, E.~P. Frady, U.~Wiklund, and E.~Osipov, ``{Density
  Encoding Enables Resource-Efficient Randomly Connected Neural Networks},''
  \emph{IEEE Transactions on Neural Networks and Learning Systems}, vol.~PP,
  no.~99, pp. 1--7, 2020.

\bibitem{scatter}
D.~Smith and P.~Stanford, ``{A Random Walk in Hamming Space},'' in
  \emph{International Joint Conference on Neural Networks (IJCNN)}, 1990, pp.
  465--470 vol.2.

\bibitem{ESN02}
H.~Jaeger, ``{Tutorial on Training Recurrent Neural Networks, Covering BPTT,
  RTRL, EKF and the Echo State Network Approach},'' Technical Report GMD Report
  159, German National Research Center for Information Technology, Tech. Rep.,
  2002.

\bibitem{FPGASTDP2019}
C.~{Lammie}, T.~J. {Hamilton}, A.~{van Schaik}, and M.~{Rahimi Azghadi},
  ``{Efficient FPGA Implementations of Pair and Triplet-Based STDP for
  Neuromorphic Architectures},'' \emph{IEEE Transactions on Circuits and
  Systems I: Regular Papers}, vol.~66, no.~4, pp. 1558--1570, 2019.

\bibitem{HACNN2018}
S.~{Moini}, B.~{Alizadeh}, M.~{Emad}, and R.~{Ebrahimpour}, ``{A
  Resource-Limited Hardware Accelerator for Convolutional Neural Networks in
  Embedded Vision Applications},'' \emph{IEEE Transactions on Circuits and
  Systems II: Express Briefs}, vol.~64, no.~10, pp. 1217--1221, 2017.

\bibitem{XPEconf}
Y.~{Eminaga}, A.~{Coskun}, and I.~{Kale}, ``{Area and Power Efficient
  Implementation of db4 Wavelet Filter Banks for ECG Applications Using
  Reconfigurable Multiplier Blocks},'' in \emph{4th International Conference on
  Frontiers of Signal Processing (ICFSP)}, 2018, pp. 65--68.

\end{thebibliography}
